\documentclass[10pt,twocolumn,letterpaper]{article}

\usepackage[pagenumbers]{cvpr} 
\usepackage{graphicx}
\usepackage{amsmath}
\usepackage{amssymb}
\usepackage{booktabs}

\usepackage{enumitem} 
\usepackage{csvsimple}

\usepackage{multirow}
\usepackage{tabularray}
\usepackage{adjustbox}

\usepackage{algorithm}
\usepackage[noend]{algpseudocode}

\usepackage[accsupp]{axessibility}  

\usepackage{titletoc}
\usepackage{tocloft}
\usepackage{etoc}

\usepackage{xcolor}

\usepackage{colortbl}

\usepackage{color, colortbl}
\definecolor{Gray}{gray}{0.9}
\definecolor{LightCyan}{rgb}{0.88,1,1}
\usepackage[first=0,last=9]{lcg}

\definecolor{LightBlue}{rgb}{0.90,0.95,1}
\definecolor{LightYellow}{rgb}{1, 1, 0.7}
\definecolor{LightGreen}{rgb}{0.7, 0.9, 0.7}

\usepackage[font=small,labelfont=bf,tableposition=top]{caption}
\usepackage{blindtext}
\usepackage{cuted}

\usepackage[pagebackref,breaklinks,colorlinks]{hyperref}

\usepackage[capitalize]{cleveref}
\crefname{section}{Sec.}{Secs.}
\Crefname{section}{Section}{Sections}
\Crefname{table}{Table}{Tables}
\crefname{table}{Tab.}{Tabs.}

\def\cvprPaperID{13142} 
\def\confName{CVPR}
\def\confYear{2024}

\begin{document}

\newcommand{\beginsupplement}{%
    \setcounter{table}{0}
    \renewcommand{\thetable}{S\arabic{table}}%
    \setcounter{figure}{0}
    \renewcommand{\thefigure}{S\arabic{figure}}%
    \setcounter{algorithm}{0}
    \renewcommand{\thealgorithm}{S\arabic{algorithm}}%
}

\title{Rotation-Agnostic Image Representation Learning for Digital Pathology}
	
\author{Saghir Alfasly \:\:  Abubakr Shafique \:\: Peyman Nejat\:\: Jibran Khan\:\: Areej Alsaafin \:\:  Ghazal Alabtah \\ H.R.Tizhoosh\thanks{Corresponding author}  \\
		\small KIMIA Lab, Department of Artificial Intelligence \& Informatics, Mayo Clinic, Rochester, MN, USA\\
        {\tt\small \{alfasly.saghir, tizhoosh.hamid\}@mayo.edu}
	}	
 
\maketitle
\begin{abstract}
This paper addresses complex challenges in histopathological image analysis through three key contributions. Firstly, it introduces a fast patch selection method, \emph{FPS}, for whole-slide image (WSI) analysis, significantly reducing computational cost while maintaining accuracy. Secondly, it presents \emph{PathDino}, a lightweight histopathology feature extractor with a minimal configuration of five Transformer blocks and only $\approx$ $9$ million parameters, markedly fewer than alternatives. Thirdly, it introduces a rotation-agnostic representation learning paradigm using self-supervised learning, effectively mitigating overfitting. We also show that our compact model outperforms existing state-of-the-art histopathology-specific vision transformers on 12 diverse datasets, including both internal datasets spanning four sites (breast, liver, skin, and colorectal) and seven public datasets (PANDA, CAMELYON16, BRACS, DigestPath, Kather, PanNuke, and WSSS4LUAD). Notably, even with a training dataset of $\approx$$6$ million histopathology patches from The Cancer Genome Atlas (TCGA), our approach demonstrates an average 8.5\% improvement in patch-level majority vote performance. These contributions provide a robust framework for enhancing image analysis in digital pathology, rigorously validated through extensive evaluation.
\footnote{The project page: \url{https://KimiaLabMayo.github.io/PathDino-Page/}}
\end{abstract}

\begin{figure}[!h] 
    \centering 
    \includegraphics[width=.30\textwidth]
    {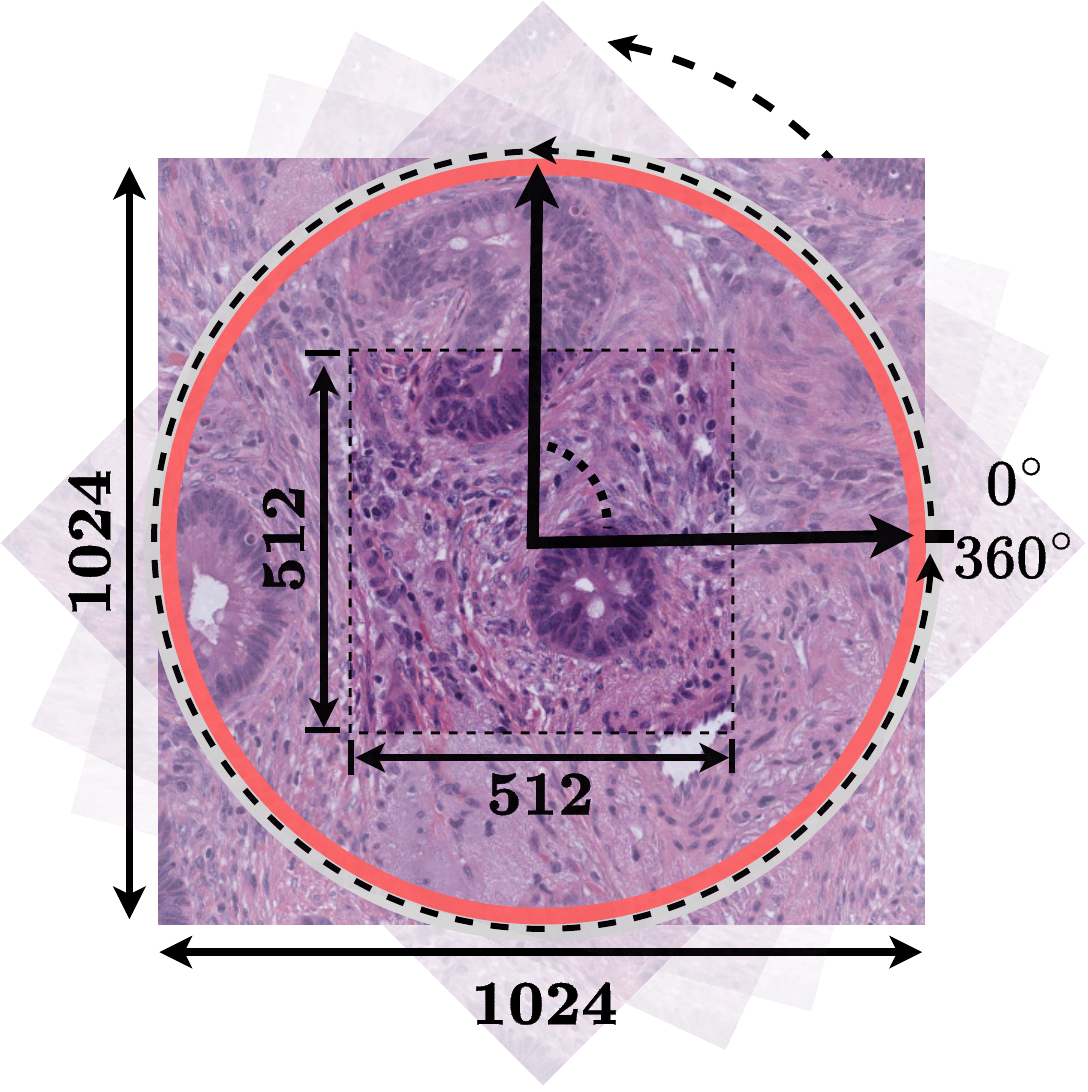}  
    \vspace{-10pt}
    \caption{\textbf{\texttt {HistoRotate.}} A $360^\circ$ rotation augmentation for training models on histopathology images. Unlike training on natural images where the rotation may change the context of the visual data, rotating a histopathology image improves the learning process for discriminative embedding learning.}  \vspace{-10pt}
    \label{fig:FigHistRotate4}
\end{figure}
\vspace{-15pt}

\section{Introduction}
\label{sec:intro}

The advent of whole slide image (WSI) scanning in digital pathology has revolutionized the research in computational pathology \cite{WSIReveiw,WSIReveiw2,WSINature}. While digital pathology enables both researchers and clinicians to enjoy the ease of access to the WSIs, processing and storing these gigapixel images are still quite challenging. 

\noindent\textbf{Motivation: }Large image size and scarce or lack of patch-level labels (annotations) pose two main challenges in WSI analysis \cite{WSIReveiw3}. As a result, most state-of-the-art methods adopt Multi-instance Learning (MIL) with weak supervision \cite{patchClustering10Clusters,patchClusteringIteratively,multiInstanceTransmil,multiInstanceDTFDMIL,multiInstanceCVPR23,multiInstancePreprintCOmparewithCLAM, multiInstancechan2023histopathology,multiInstanceli2023task,BFpatchingchen2021multimodal}. While these approaches may eliminate the need for pixel-level annotations, MIL significantly increases computational loads and potentially lowers the quality of results compared to fully supervised approaches. While some attempts have been made to select representative patches \cite{yottixel, patchClustering10Clusters, patchClusteringIteratively, patchIdentifyThenSelect}, many such methods remain computationally intensive, leaving the desire for efficient, accurate solutions an unmet need.

The field of image analysis in digital pathology has predominantly adopted deep models designed for natural image analysis without further customization to the field \cite{TransPath,HIPT,SSLPath,iBOT}. While showing good performance on natural image analysis, pre-trained deep models may not fully exploit the unique characteristics of histopathology images. Furthermore, most current training recipes for histopathological embedding learning adopt conventional training and common augmentation techniques for natural images \cite{SSLPath}. However, histopathology images have arguably very different features compared to natural images and even radiology images. This gap motivated us to design an improved training approach for histopathology images.

\noindent\textbf{Contributions:} We present a two-fold solution that encompasses selective patching and robust feature extraction. First, we propose a fast patch selection \textbf{\texttt {FPS,}} a ``\emph{divide \& conquer}'' algorithm that is capable of identifying a compact and yet highly representative subset of patches for analysis. This algorithm has been meticulously tuned to balance computational efficiency and diagnostic utility. Secondly, we introduce \textbf{\texttt {PathDino}} a lightweight histopathology-specific transformer consisting of just five small vision transformer blocks, customized and  finely tuned to the nuances of histopathological images. It not only exhibits superior performance but also effectively reduces susceptibility to overfitting. We also propose \textbf{\texttt{HistoRotate}}, a seamless $360^\circ$ rotation augmentation technique designed specifically for training histopathology  models. The incorporation of this augmentation technique with the proposed lightweight histopathology-specific transformer results in a significant enhancement of embedding quality and effectively mitigates overfitting. Our model is rigorously validated through extensive evaluation on multiple datasets, showing both computational efficiency and  superior performance. Overall, our \textbf{key contributions} are as follows:	

\begin{itemize}
\item \textbf{Fast Patch Selection}: A novel and efficient patch selection mechanism curates a compact, spatially diverse subset of patches from WSI, reducing computational overhead while maintaining representational fidelity.
\vspace{-5pt}
\item \textbf{PathDino}: A lightweight histopathology-specific Vision Transformer with only $5$ transformer blocks, totaling $9$ million parameters, offering reduced susceptibility to overfitting.
\vspace{-5pt}
\item \textbf{Rotation-Agnostic Representation Training}: We propose \textbf{\texttt {HistoRotate}}, a $360^\circ$ rotation augmentation technique designed for training histopathological image analysis models. Unlike natural images, rotating histopathological patches maintain the general context while enhancing embedding learning for improved reliability.
\vspace{-5pt}
\item \textbf{Extensive Evaluation}: Rigorous validation through comprehensive experiments across eleven datasets, demonstrating competitive to superior performance compared to existing state-of-the-art methods.
\end{itemize}
\vspace{-5pt}

\section{Related Work}

\noindent\textbf{WSI Patching.} WSI patching is a fundamental phase in WSI analysis pipelines, although it has received limited attention in the field. Many methods employ a brute force tiling approach, where the entire WSI is divided into thousands of patches \cite{multiInstanceCLAM,multiInstanceTransmil,BFpatchingguan2022node,multiInstanceli2021dual}, typically utilized with weakly supervised training methods like multi-instance learning \cite{patchClustering10Clusters,patchClusteringIteratively,multiInstanceTransmil,multiInstanceDTFDMIL,multiInstanceCVPR23,multiInstancePreprintCOmparewithCLAM, multiInstancechan2023histopathology,multiInstanceli2023task,BFpatchingchen2021multimodal}. This approach is often employed when only WSI-level labels are available, as in TCGA, instead of pixel-level annotations \cite{DataPANDA,DataBRACS, DataCAMELYON16}. However, brute force patch processing proves very challenging in practice due to the immense computational costs and potential training instability.

\noindent\textbf{Clustering-Based Patch Selection.} This approach aims to address patch quality by selecting representative patches but introduces new degrees of freedom such as number of clusters. It includes both \emph{Independent Patching Phase}, where only one method in the literature, namely Yottixel's mosaic \cite{yottixel}, follows this independent approach. Yottixel employs a two-stage clustering process, first based on color (stain) features and then on connected regions, creating a patch set with visual and spatial diversity. At the end, it uses a guided sampling inside each cluster. It stands as the only independent patching method adaptable to various WSI analysis pipelines. In contrast, the \emph{Integrated Patching Phase} tightly couples patching methods with specific WSI analysis methods, limiting their applicability to other uses. For example, in \cite{patchClustering10Clusters}, patch clustering is performed for each WSI into $k$ clusters, integrated with Multi-instance learning. Similarly, in \cite{patchClusteringIteratively}, a similar approach is used, clustering the entire dataset patches into a few clusters and matching specific WSI patches with cluster centroids, effectively assigning patches with pseudo labels. 

While embedded clustering methods prove inflexible and unsuitable for integration into other WSI pipelines, approaches based on clustering, although enhancing the quality of the chosen patch set, concurrently introduce an additional layer of parameters and variability to the overall process. 
To address these challenges, we propose a new fast patch selection method that avoids the brute-force and multi-variable clustering approaches. Crucially, our \textbf{\texttt {FPS}} aligns with the independent patching phase, exemplified by Yottixel, enhancing adaptability for WSI analysis pipelines while greatly improving efficiency.

\noindent\textbf{Vision Transformer in Histopathology.} A prevalent trend in histopathological image analysis is the adaptation of mainstream vision transformers, especially ViT (Vision Transformer) \cite{ViT, liu2021swin}. Many existing models are essentially fine-tuned versions of ViT \cite{TransPath,HIPT, SSLPath,iBOT}, often overlooking the unique characteristics of histopathological images compared to natural images, leading to issues such as overfitting since ViTs are known to be data-hungry \cite{dosovitskiy2020image}. In contrast, our comparably compact ViT architecture \textbf{\texttt {PathDino}} tailored for histopathological images, achieving better results while mitigating overfitting.

\noindent\textbf{Self-Supervised Learning in Digital Pathology.} Self-supervised learning has gained popularity in digital pathology due to its independence from annotated histopathological images, making it possible to leverage large datasets \cite{DinoV1, DinoV2}. However, most self-supervised learning approaches are primarily developed for natural image analysis \cite{SwAV, MoCoV2, DinoV1, DinoV2, MIMiBOT, BarlowTwins}. Applying these methods directly to histopathological embedding learning without considering domain-specific differences can lead to suboptimal performance. Recent studies underscore the value of domain-specific pre-training for transferability. Domain-specific self-supervised learning methods are also shown to significantly enhance performance in medical imaging tasks \cite{39matsoukas2022makes, 2azizi2021big, 3boyd2021self, 16ciga2022self, 23gildenblat2019self, 48sowrirajan2021moco, TransPath, 60yang2022concl}. Furthermore, BYOL, SimSiam, and SimCLR frameworks have been employed for image classification and patch retrieval in histopathology \cite{16ciga2022self, 23gildenblat2019self, TransPath, multiInstanceli2021dual}.

Recent studies have shown promising results in enhancing model performance for downstream tasks in medical imaging through transfer learning and domain-specific self-supervised learning methods. Kang \etal. in \cite{SSLPath} conducted a comprehensive benchmarking study on self-supervised representation learning in histopathology images, evaluating several methods on a dataset of $32.6$M patches ($19$M from TCGA\footnote{https://portal.gdc.cancer.gov/} and $13.6$M from TULIP which is an private dataset), including SwAV, MoCoV2, Barlow Twins, and DinoV1 \cite{SSLPath}. Hierarchical Image Pyramid Transformer (HIPT) is a self-supervised Transformer trained on TCGA patches using Dino-based self-supervised training, whereas TransPath is a self-supervised model trained on TCGA and PAIP patches through contrastive learning \cite{HIPT, TransPath}. iBOT-Path \cite{iBOT}, a vision transformer, was trained on $40$M histopathology patches from TCGA using the self-supervised iBOT framework \cite{MIMiBOT}. Additionally, models like BiomedCLIP \cite{BiomedGPT} and PLIP \cite{PLIP}, trained with image-text contrastive learning on the biomedical PMC-15M dataset and the histopathology dataset OpenPath, respectively. Virchow \cite{vorontsov2023virchow}, a Transformer-based model with $632$ million parameters, was trained using DinoV2-based self-supervised learning on $1.5$M internal WSIs \cite{DinoV2}.

\noindent\textbf{Our work differs from previous methods in the following aspects:} \textbf{\emph{WSI Patching}}: Our FPS method offers superior efficiency compared to \cite{yottixel} without the need for patch clustering, while still maintaining competitive accuracy. \textbf{\emph{Histopathology-specific ViT Structure}}: Our PathDino is a lightweight ViT that contains only 5 small transformer blocks for effective histopathological image analysis. \textbf{\emph{Training Recipe}}: Our training recipe features  \emph{HistoRotate} augmentation that applies $360^\circ$ rotation leading to rotation-invariant embedding learning. 

\begin{figure*}
    \centering 
    \includegraphics[width=\textwidth]{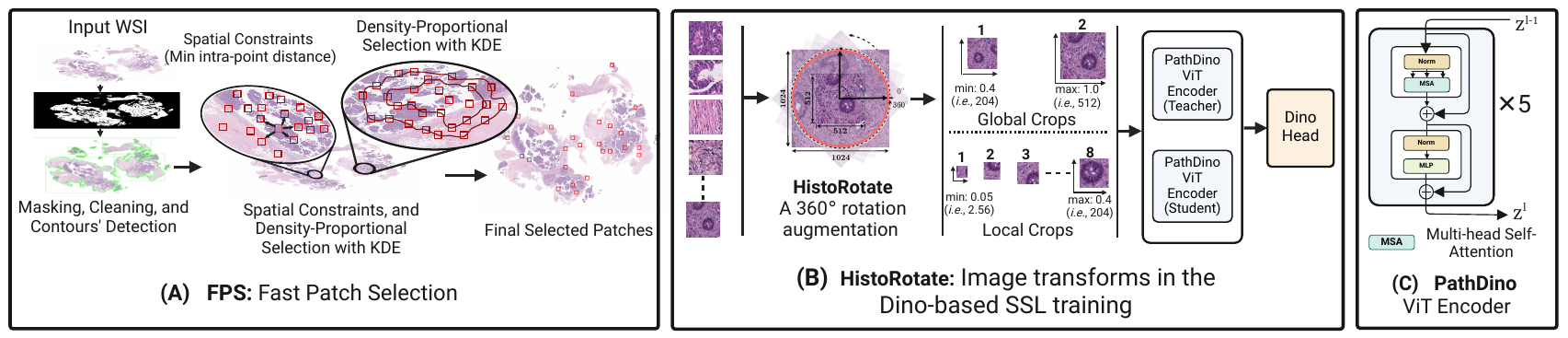} \vspace{-5pt} 
    \caption{\textbf{The WSI Analysis Pipeline.} (A) The fast patch selection method, FPS, selects a set of representative patches while preserving spatial distribution. (B) HistoRotate is a $360^\circ$ rotation augmentation for histopathology model training, enhancing learning without contextual information alteration. (C) PathDino is a compact histopathology Transformer with five small vision transformer blocks and $\approx$$9$ million parameters, significantly leaner than alternatives.}  
    \label{fig:fullPipeline}
\end{figure*}

\section{Proposed Method}
\subsection{FPS: Fast Patch Selection}
In this section, we introduce a method for the systematic selection of representative patches from WSIs for computational pathology. The algorithm aims to cater to both the diversity and relevance of the tissue structure, thus capturing the inherent complexity and heterogeneity of tissue slides as illustrated in Fig. \ref{fig:fullPipeline}-A.

\noindent\textbf{Preprocessing.} Given a WSI, \(I\), with dimensions \(W \times H\), a thumbnail image, \(T\), with dimensions \(w \times h\) is generated. A tissue mask, \(M\), is obtained through binary thresholding.

\noindent\textbf{Density-Proportional Selection with Kernel Density Estimation (KDE)}
The contours extracted \cite{FindContourssuzuki1985topological} from the tissue mask are denoted by \(C\), where \(C = \{c_1, c_2, \ldots, c_n\}\). For each contour \(c_i\), a bounding box is defined as \(R_i = [x, y, w, h]\). A set of potential patch locations, \(P\), is constructed as follows:
\begin{equation}
    \begin{gathered}
    P = \bigcup_{i=1}^{n} \{(x, y) \mid x \in [R_{i,x}, R_{i,x} + R_{i,w} - r_w], \\
    y \in [R_{i,y}, R_{i,y} + R_{i,h} - r_h]\},
    \end{gathered}
    \label{eq:potential_patches}
\end{equation}
where, \(r_w\) and \(r_h\) are the dimensions of the patches in the mask space. Subsequently, density-proportional KDE is employed to generate the set \(S\) of selected patches:
\vspace{-5pt}
\begin{equation}
    S = \text{KDE}(P, n_s),
    \label{eq:patch_selection}
\end{equation}
where \(n_s\) is the predefined number of patches to be selected.
Utilizing the KDE to approximate the probability density function \(f(x)\) over the set \(P\) is performed as follows:
\begin{equation}
    f(x) = \frac{1}{Nh} \sum_{i=1}^{N} K\left(\frac{x-x_i}{h}\right),
    \label{eq:kde_estimation}
\end{equation}
where \(K\) is the kernel function, \(N\) is the total number of points in \(P\), and \(h\) is the bandwidth (\ie, the width of the smoothing kernel).

\noindent\textbf{Density-Proportional Sampling.}
In accordance with the density map generated by KDE, points are sampled proportionally to their density values:
\vspace{-5pt}
\begin{equation}
    p(x) = \frac{f(x)}{\sum_{x \in P} f(x)}.
    \label{eq:prob_selection}
\end{equation}

A random sample \(S\) consisting of \(n_s\) points is extracted from \(P\) based on the probability density function \(p(x)\):
\vspace{-5pt}
\begin{equation}
    S = \text{Rand}(P, p(x), n_s).
    \label{eq:final_sample}
\end{equation}

The resulting set \(S\) conforms to the spatial density characteristics of the tissue structures in the slide, thus capturing the tissue heterogeneity.

\noindent\textbf{Spatial Constraints.}
To avoid oversampling from densely packed regions, a minimum Euclidean distance, \(e_{\text{min}}\), is enforced between any two selected patches \(s_i\) and \(s_j\):
\vspace{-5pt}
\begin{equation}
    \forall s_i, s_j \in S, \sqrt{(s_{i,x} - s_{j,x})^2 + (s_{i,y} - s_{j,y})^2} \geq e_{\text{min}}.
    \label{eq:spatial_constraint}
\end{equation}\vspace{-5pt}


Finally, the selected patches are mapped back to the WSI coordinates at high magnification for downstream analyses. Each patch location \((x, y) \in S\) is scaled to its corresponding location in \(I\) using the ratio between \(W\) and \(w\), as well as \(H\) and \(h\). The patches are extracted and stored for subsequent analyses.

\subsection{HistoRotate: Rotation-Agnostic Training}

In addressing the unique challenges in tissue image analysis, we introduce a new self-supervised training recipe that incorporates a rotation-agnostic scheme as depicted in Fig. \ref{fig:FigHistRotate4}, designed to enhance the quality of the learned representations by incorporating various angular rotations of the image during the training process. Let \( I \) denote an input image, and \( \theta \) denote a randomly selected angle from a predefined set \( \Theta \). The rotation operation \( \mathcal{R} \) is formally defined as:
\begin{equation}
\mathcal{R}(I, \theta) = I_\theta.
\end{equation}
In our implementation, two types of rotations are considered:

\noindent(a) \emph{Random Continuous Rotation}: \( \theta \) is sampled from a continuous uniform distribution over the range \([0, 360]\) degrees.
\begin{equation}
    \theta \sim \mathcal{U}(0, 360).
\end{equation}
\noindent(b) \emph{Random Discrete Rotation}: \( \theta \) is selected from the set \( \Theta = \{90, 180, 270, 360\} \).
Each image undergoes a cropping operation \( \mathcal{C} \) before and after the rotation, followed by a resizing operation \( \mathcal{S} \) to generate a transformed image \( I' \).
\begin{equation}
I' = \mathcal{S}(\mathcal{C}(\mathcal{R}(I, \theta))).
\end{equation}
\noindent\textbf{\emph{HistoRotate} with Dino Framework}. As depicted in Fig. \ref{fig:fullPipeline}-B, we applied these transformations on two types of image crops used in Dino framework \cite{DinoV1}:
\textbf{Global Crops}: Images are cropped and resized to a scale \( s \) sampled from \( \mathcal{U}(0.4, 1) \).
\textbf{Local Crops}: Images are cropped and resized to a smaller scale \( s' \) sampled from \( \mathcal{U}(0.05, 0.4) \).
In the final data augmentation pipeline, we generate a set \( \mathcal{I} \) of transformed images from each original image \( I \):
\begin{equation}
\mathcal{I} = \{ I'_1, I'_2, \ldots, I'_n \}.
\end{equation}

The proposed rotation-agnostic representation learning scheme yields a significant advantage in obtaining more comprehensive and robust tissue image representations.

\subsection{PathDino: A Histopathology-specific Vision Transformer}
\begin{figure}
    \centering 
    \includegraphics[width=\columnwidth]{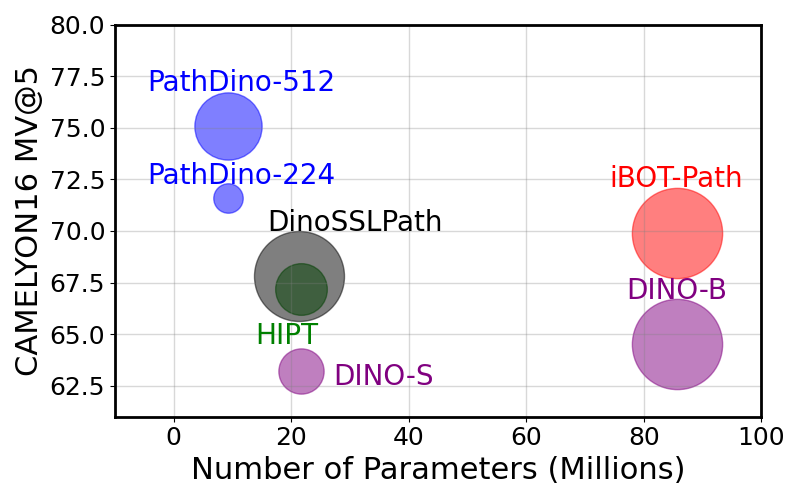}  \vspace{-7pt}
    \caption{\textbf{PathDino vs. its counterparts.} Number of parameters (millions) vs. the patch-level retrieval with macro average $F$-$1$ score of majority vote (MV@5) on CAMELYON16 dataset. The bubble size represents the FLOPs. }  
    \label{fig:FLOPs}
\end{figure}

We introduce PathDino, a shallow and compact vision transformer designed for histopathological image analysis. This model is lightweight and less prone to overfitting. It has an embedding size of $d=384$, $6$ attention heads, and a patch size of $16\times 16$ for input images $X \in \mathbb{R}^{H\times W \times C}$. We evaluate two input resolutions: $H=W=512$ (PathDino-$512$) and $H=W=224$ (PathDino-$224$). PathDino encoder comprises a total of $L=5$ blocks. Each block consists of a multi-head self-attention (MSA) layer, LayerNorm (LN), and a multilayer perceptron (MLP):
\begin{equation}
\mathbf{z}_i^{\ell} = \mathrm{MLP}(\mathrm{LN}(\mathrm{MSA}(\mathbf{z}_i^{\ell-1}))) + \mathbf{z}_i^{\ell-1},
\end{equation}
where  $\mathbf{z}_i \in \mathbb{R}^d$, $\quad \ell=1,\cdots,L$, and $i=1,\cdots,N$ and $N$ here represents the total input transformer patches. Fig. \ref{fig:fullPipeline}-C visualizes PathDino encoder structure, whereas Fig. \ref{fig:FLOPs} visually compares PathDino's performance, FLOPs, and parameter count with those of its counterparts. PathDino contains $\approx$$9$M parameters, significantly fewer than ViT-s ($21$M) used by DinoSSLPath \cite{SSLPath} and HIPT \cite{HIPT}, as well as the ViT-b ($85$M) used by iBOT-Path \cite{iBOT}.

\section{Experiment Setup}
\noindent\textbf{Hardware:} All experiments have been conducted on a Dell PowerEdge XE8545 server with 4$\times$ NVIDIA A100-SXM4-80GB  and 2$\times$ AMD EPYC 7413 CPUs, 1023 GB RAM.
\noindent\textbf{PathDino Pretraining Dataset}. We extracted a total of \(6,087,558\) patches from \(11,765\) diagnostic TCGA WSIs. Specifically, \(3,969,490\) patches have a \(1024 \times 1024\) dimension, while \(2,118,068\) patches have a \(512 \times 512\) dimension. The extraction was conducted at a \(20\times\) magnification level, with a patch tissue area threshold of \(90\%\).

\noindent\textbf{PathDino Pretraining Details.} All pretraining and evaluation processes are conducted using the \emph{Pytorch} deep learning library and Python. We adapt DINO \cite{DinoV1} framework in which we integrated our augmentation method \emph{HistoRotate} to be applied to each cropped image portion of the internal and global crops. In the pretraining phase of our study, we utilized $\approx6$M patches from TCGA. To ensure high-quality data selection, a tissue threshold of $90\%$ was employed to filter the patches without enough tissue coverage from the WSIs. Our pretraining approach follows self-supervised learning, implemented on top of the DINO framework. We employed two sets of crops, comprising $2$ global crops and $8$ local crops. Our pretraining efforts resulted in the development of two distinct models: PathDino-224, trained on $224\times 224$ cropped images obtained solely from the $2,118,068$ patches with size $512\times 512$. We utilized a batch size of $384$ with the AdamW optimizer and a learning rate of $0.0001$ for $30$ epochs. Meanwhile, PathDino-512, a model with $512\times 512$ dimensions trained on the entire $6,087,558$ patches for $27$ epochs employing a batch size of $192$ and the AdamW optimizer with an initial learning rate of $0.0005$.

\noindent\textbf{Downstream Datasets}.
\textbf{\emph{Private Skin:}} Contains $660$ WSIs primarily capturing cutaneous squamous cell carcinoma (cSCC) biopsies in various differentiation stages including a class of normal skin biopsy. Demographic features indicate a median patient age of $77$, with females making up $35\%$ of the dataset.
\textbf{\emph{Private Liver:}} Includes $150$ WSIs of alcoholic steatohepatitis (ASH), $158$ WSIs of non-alcoholic steatohepatitis (NASH), and $18$ WSIs of normal cases predominantly sourced from liver biopsies.
\textbf{\emph{Private CRC:}} Features $209$ WSIs, categorized into Cancer Adjacent Polyp (CAP), Non-recurrent Polyp (POP-NR), and Recurrent Polyp (POP-R) classes.
\textbf{\emph{Private Breast:}} Consists of $73$ WSIs classified into $16$ tumor subtypes and one class of normal tissue, encapsulating a variety of pathological conditions such as Adenoid Cystic Carcinoma (ACC), Ductal Carcinoma In Situ (DCIS), among others.
\textbf{\emph{PANDA \cite{DataPANDA}:}} A public dataset of $12,625$ WSIs of prostate biopsies stained with H\&E, collected from diverse international sites for comprehensive evaluation.
\textbf{\emph{CAMELYON16 \cite{DataCAMELYON16}:}} Provides $399$ meticulously annotated WSIs of lymph node sections collected from breast cancer patients across two hospitals in the Netherlands.
\textbf{\emph{BRACS \cite{DataBRACS}:}} Encompasses $547$ WSIs from $189$ patients, annotated into seven distinct lesion subtypes by board-certified pathologists.
\textbf{\emph{DigestPath \cite{DataDigestPath}:}} Comprises two specialized datasets for diagnosing gastrointestinal histopathology features: the Signet Ring Cell Detection Dataset (SRC) and the Colonoscopy Tissue Segmentation and Classification Dataset (TSCC).
\textbf{\emph{PanNuke \cite{DataPanNuke}:}} A semi-automatically generated nuclei instance segmentation and classification dataset containing exhaustive nuclei labels across $19$ different tissue types.
\textbf{\emph{Kather-7K \cite{DatasetKather}:}} Features $7,180$ non-overlapping image patches sourced from $50$ patients with colorectal adenocarcinoma, serving as an ideal validation set for model evaluation.
\textbf{\emph{WSSS4LUAD \cite{DataWSSS4LUAD}:}} Specifically built for segmentation tasks in lung adenocarcinoma histopathology, including over $10,091$ patch-level annotations.
Additional details for each dataset are available in the Suppl-Tables [\ref{tab:privateDataset}, \ref{tab:publicDataset}].

\noindent\textbf{Evaluation Metrics.}
For the evaluation of WSI-level and patch-level retrievals, we used Top-1, the majority vote among Top-3 (MV@3), and the majority vote among Top-5 (MV@5) metrics within the leave-one-out evaluation scheme. To assess the patch classification task, we trained a linear classifier using the extracted feature embeddings and computed accuracy and macro average $F$-$1$ score. Embedding variances were analyzed using Principal Component Analysis, as illustrated in Figure \ref{fig:FigEmbeddingVariance}. Additionally, the quality of the Vision Transformer (ViT) is visually assessed using activation maps, as shown in Figure \ref{fig:FigAttenCompare3Models}. An extensive evaluation, both qualitative and quantitative, is presented in the subsequent sections and the Suppl. File.\vspace{-3pt}

\section{Experimental Results}
\label{sec:results}

\subsection{FPS Effectiveness}
\label{sec:patchingEffectiveness}
Table \ref{tab:wsiLevelFPSvsYottixel} provides an in-depth comparative assessment between Yottixel's mosaic and our FPS patching method across 3 private and 3 public histopathology datasets, utilizing BiomedCLIP \cite{zhang2023large}, PLIP \cite{PLIP}, and PathDino as backbones. Across internal datasets, FPS consistently exhibits competitive to superior performance. For example, in Private-Breast dataset, FPS achieves a top-1 accuracy of 58\% with PLIP and 68\% with PathDino, outperforming Yottixel's corresponding values of 55\% and 63\%. In Private-Liver dataset, FPS integrated with PathDino achieves an 83\% top-1 accuracy, markedly higher than Yottixel's accuracy of 81\%. This trend is corroborated in the Private-Skin and Private-CRC datasets, where FPS surpasses Yottixel's mosaic in all metrics, most notably achieving an MV@5 of 82\% in Private-Skin with PLIP, but lower performance on Top1 and MV@3. The results in public datasets demonstrate on par performance rather than superiority. For example, in the PANDA dataset, FPS, when paired with PLIP, records a top-1 accuracy of 56\%, which is 3\% higher than Yottixel's mosaic. In summary, the empirical evidence overwhelmingly supports the efficacy of FPS as compared to Yottixel's mosaic. More results are reported in Suppl-Tables [\ref{tab:wsiLevelFPSvsYottixelAcc}, \ref{tab:wsiLevelFPSvsYottixelF1}].

\begin{figure}
    \centering 
    \includegraphics[width=.47\textwidth]{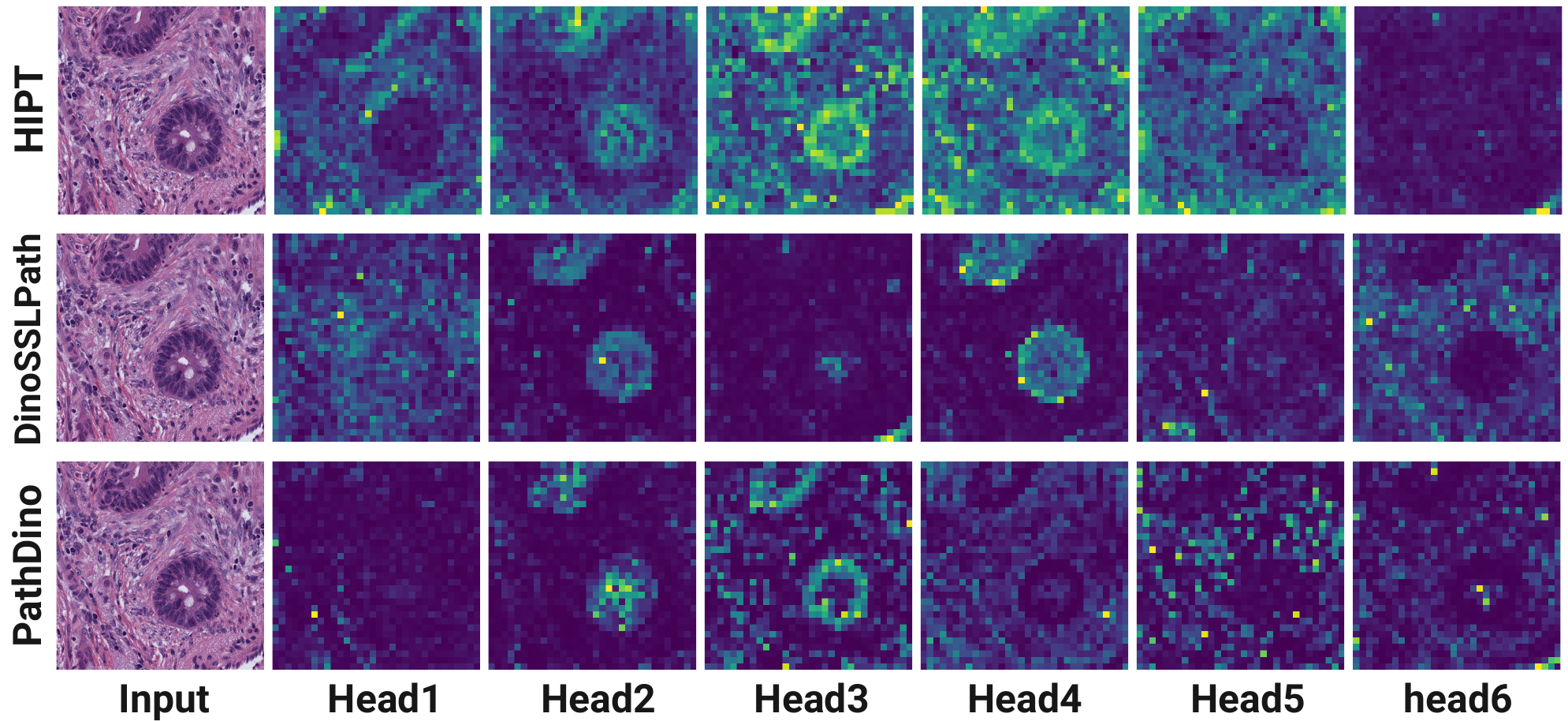}   \vspace{-5pt}
    \caption{\textbf{Attention Visualization.} When visualizing attention maps, our PathDino transformer outperforms HIPT-small and DinoSSLPath, despite being trained on a smaller dataset of $6$M TCGA patches. In contrast, DinoSSLPath and HIPT were trained on much larger datasets, with $19$ million and $104$ million TCGA patches, respectively.} 
    \label{fig:FigAttenCompare3Models}
\end{figure}

\begin{table}
    \centering
\caption{Performance accuracy of the proposed FPS against Yottixel's mosaic using BiomedCLIP, PLIP and PathDino backbones.} \vspace{-5pt}
    \begin{adjustbox}{width=\columnwidth}
    \small
    \begin{tabular}{clccc|cc|cc}
    \hline
    & \multirow{2}{*}{Dataset}&& \multicolumn{2}{c|}{BiomedCLIP \cite{zhang2023large}}& \multicolumn{2}{c|}{PLIP \cite{PLIP}} & \multicolumn{2}{c}{PathDino}\\
    \cline{4-9}
    &&& Yottixel & FPS &Yottixel & FPS& Yottixel & FPS \\  \hline 
    \multirow{10}{*}{\rotatebox[origin=c]{90}{Internal Data}} 
    & \multirow{1}{*}{Private-Breast} & Top 1 &\textbf{47}&\textbf{47}&55& \textbf{58}&63& \textbf{68}\\
    \cline{2-9}
    
    & \multirow{3}{*}{Private-Liver} & Top 1 &70&\textbf{74}& 70 & \textbf{73}&81& \textbf{83} \\
    & & MV@3 &75&\textbf{77}& \textbf{76} &74 &86&\textbf{86}\\
    & & MV@5 &74&\textbf{77}& 73 & \textbf{76}&\textbf{87}&85\\
    \cline{2-9}
    
    & \multirow{3}{*}{Private-Skin} & Top 1 &68&\textbf{75}& 72 & \textbf{75}&\textbf{79}&78\\
    & & MV@3 &73&\textbf{78}& 77 &\textbf{79} & \textbf{81}&80\\
    & & MV@5 &76&\textbf{78}& 80 &\textbf{82} &81&\textbf{82}\\
    \cline{2-9}
    
    & \multirow{3}{*}{Private-CRC} & Top 1 &55&\textbf{58}& 60 & \textbf{64}&57& \textbf{63}\\
    & & MV@3 &60&\textbf{63}& 61 &\textbf{67}&60&\textbf{65}\\
    & & MV@5 &59&\textbf{65}& 62 & \textbf{69}&61&\textbf{65}\\
    \cline{1-9}\cline{1-9}
    
    \multirow{9}{*}{\rotatebox[origin=c]{90}{Public Data}} & \multirow{3}{*}{PANDA \cite{DataPANDA}} & Top 1 &33&\textbf{34}&53 & \textbf{56}&\textbf{59}&58\\
    & & MV@3 &\textbf{36}&\textbf{36}& 53 & \textbf{55}&58&\textbf{58}\\
    & & MV@5 &\textbf{38}&\textbf{38}& 53 & \textbf{54}&\textbf{58}&56\\
    \cline{2-9}
    
    & \multirow{3}{*}{CAMELYON16 \cite{DataCAMELYON16}} & Top 1 &60&\textbf{61}& 70 &\textbf{73} &\textbf{76}& 73\\
    & & MV@3 &58&\textbf{67}& 71 &\textbf{77} &77& \textbf{78}\\
    & & MV@5 &64&\textbf{69}& 70 &\textbf{75} &\textbf{78}&77\\
    \cline{2-9}
    
    & \multirow{3}{*}{BRACS \cite{DataBRACS}} & Top 1 &\textbf{56}&55& \textbf{62} &60 &\textbf{65}&64\\
    & & MV@3 &58&\textbf{62}& \textbf{64} &63 &65&\textbf{66}\\
    & & MV@5 &59&\textbf{61}& \textbf{66} &64 &66&\textbf{67}\\
    \hline
    \end{tabular}
    \end{adjustbox}
    \label{tab:wsiLevelFPSvsYottixel}
\end{table}

\subsection{FPS Efficiency}
Table \ref{tab:propertiesOnly} elucidates the computational efficiency and processing capabilities of both patching methods when paired with the PathDino backbone. Remarkably, FPS demonstrates higher computational efficiency in most scenarios. For instance, FPS processes Private-Breast and Private-Skin datasets in significantly less time, requiring only $13.1$ and $132.0$ minutes in total, respectively, as opposed to Yottixel's $20.4$ and $171.3$ minutes. Additionally, FPS succeeds in processing more WSIs with fewer failures; in the PANDA dataset, FPS processes $138$ missed WSIs compared to Yottixel's $268$. This efficiency extends to other datasets, such as Private-CRC and BRACS, where FPS outperforms Yottixel's mosaic in both speed and the number of processed WSIs. These empirical findings not only validate the robustness and efficacy of FPS but also its computational advantages, underscoring its suitability for large-scale, time-sensitive histopathological image analysis.

\begin{table}
\centering
\caption{Comparison of FPS against Yottixel's mosaic in terms of the dataset properties such as number of extracted patches and average processing speed. For fair comparison, both frameworks use PathDino as the backbone.} \vspace{-5pt}
    \begin{adjustbox}{width=\columnwidth}
\begin{tabular}{lrrr|rr|cc}
\hline
\multirow{2}{*}{Dataset} & \multirow{2}{*}{\# WSI} & \multicolumn{2}{c|}{Extracted Patches} & \multicolumn{2}{c|}{Patching Speed (m)}& \multicolumn{2}{c}{\# missed WSI}\\\cline{3-8}
& & Yottixel & FPS & Yottixel & FPS& Yottixel & FPS\\\cline{1-8}
Private-Breast  &74& $1,141$ & $2,033$ & 20.4& \textbf{13.1}&\textbf{1}&\textbf{1}\\
Private-Liver   &326& $2,974$ & $8,297$ & \textbf{45.4}& 64.6&\textbf{2}&3\\
Private-Skin    &660& $8,388$ & $16,491$ &171.3 & \textbf{132.0}&1&\textbf{0}\\
Private-CRC     &209& $4,619$ & $6,068$ & 79.4&\textbf{46.0} &\textbf{0}&\textbf{0}\\\hline\hline

PANDA        &10617& $87,451$ & $112,763$ & 251.9&\textbf{192.1} &268&\textbf{138}\\
CAMELYON16   &129& $2,864$  & $3,870$ & 84.5& \textbf{12.9}&1&\textbf{0}\\
BRACS        &547& $12,946$ & $15,352$ &261.5 &\textbf{117.7} &24&\textbf{12}\\
\hline
\end{tabular}
    \end{adjustbox}
\label{tab:propertiesOnly}
\end{table}

\begin{figure}
    \centering 
    \includegraphics[width=\columnwidth]{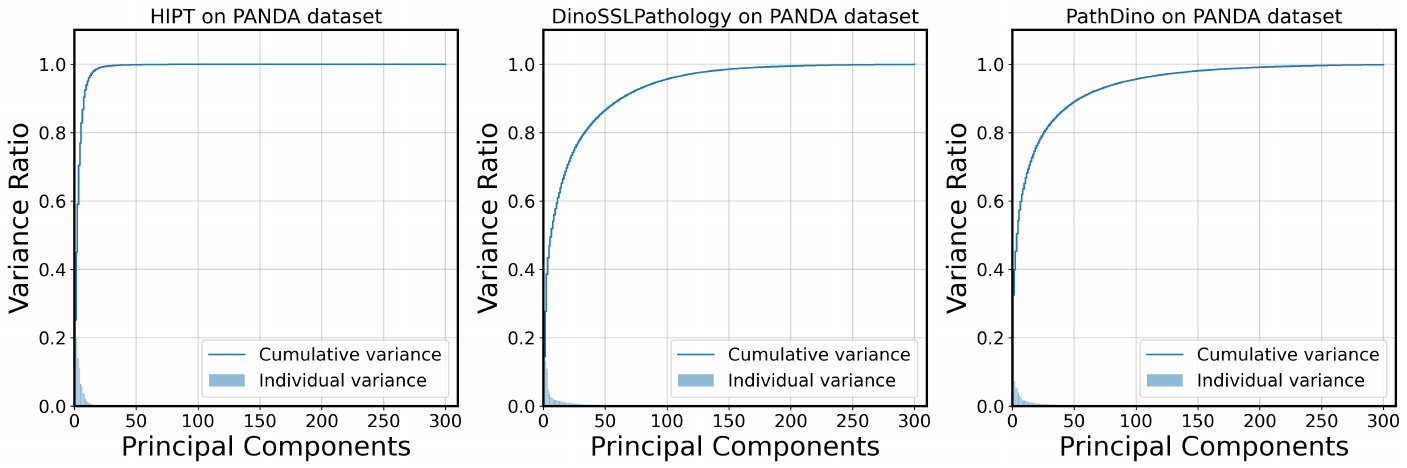}   \vspace{-15pt}
    \caption{Embedding variance analysis of three selected Transformer-based histopathological feature extractors with the output vector size of $384$ including HIPT, DinoSSLPath, and our PathDino on PANDA dataset \cite{DataPANDA}.   }  
    \label{fig:FigEmbeddingVariance}
\end{figure}

\begin{table}
    \centering
\caption{WSI-level top-1 accuracy using the proposed FPS patching method and ``\emph{median of minimum}'' Euclidean distances as proposed in Yottixel \cite{yottixel}.} \vspace{-5pt}
    \begin{adjustbox}{width=\columnwidth}
    \small
    \begin{tabular}{lccc|ccc}
    \hline%
    & \multicolumn{3}{c|}{Internal Datasets} & \multicolumn{3}{c}{Public Datasets} \\\cline{1-7}
    & Breast & Liver &  Skin & PANDA& CAMELYON16& BRACS \\
    \hline
    ResNet50 \cite{Resnet}&0.48&0.67&0.73&0.32&0.54&0.53  \\ 
    DenseNet121 \cite{DenseNet}&  0.48&0.64&0.69&0.30&0.67&0.52   \\ 
    EfficientNet-b3-288 \cite{Efficientnet}&  0.41&0.66&0.73&0.32&0.59&0.55  \\ 
    EfficientNet-b5 \cite{Efficientnet} &  0.51&0.71&0.71&0.37&0.57&0.54   \\ 
    ConvNext-b-224 \cite{Convnext}& 0.56&0.75&0.74&0.34&0.62&0.58     \\ 
    ConvNext-xlarge  \cite{Convnext}&  0.56&0.76&0.74&0.35&0.61&0.58     \\\cline{1-7}
    ViT-b16-224 \cite{ViT}& 0.41&0.7&0.72&0.31&0.6&0.54   \\ 
    DinoV1-ViT-s16 \cite{DinoV1}&  0.48&0.71&0.74&0.36&0.67&0.6     \\
    DinoV1-ViT-b16  \cite{DinoV1}&  0.55&0.72&0.73&0.37&0.63&0.59    \\
    DinoV2-ViT-b14  \cite{DinoV2}&  0.53&0.71&0.72&0.31&0.61&0.51     \\
    CLIP - ViT-B/16 \cite{CLIP}& 0.49&0.67&0.75&0.36&0.67&0.58     \\\hline
     MuDiPath-ResNet50 \cite{MuDiPath}& 0.44&0.7&0.72&0.35&0.63&0.51    \\ 
     MuDiPath-DenseNet-101  \cite{MuDiPath}&  0.51&0.68&0.74&0.36&0.65&0.56  \\
     KimiaNet \cite{kimianet} &  0.51&0.78&0.75&0.57&\textbf{0.76}&0.62    \\\cline{1-7}
     BiomedCLIP - \cite{zhang2023large} & 0.47&0.74&0.75&0.34&0.61&0.55     \\
     HIPT-ViT-s16 \cite{HIPT} &  0.44&0.68&0.73&0.32&0.62&0.52    \\ 
     PLIP \cite{PLIP} &  0.58&0.73&0.75&0.56&0.73&0.60    \\ 
     iBOT-Path  \cite{iBOT} & 0.64&0.79&0.76&0.53&0.67&\textbf{0.64}     \\ 
     DinoSSLPathology-8 \cite{SSLPath} &  0.58&0.74&\textbf{0.78}&0.47&0.74&0.61  \\\hline \rowcolor{LightBlue}\hline
     PathDino-224 (ours)&  0.53&0.75&0.74&0.46&0.72&0.61   \\ \rowcolor{LightBlue}
     PathDino-512 (ours)& \textbf{0.68}&\textbf{0.83}&\textbf{0.78}&\textbf{0.58}&0.73&\textbf{0.64}   \\\hline
    \end{tabular}
    \end{adjustbox}
    \label{tab:wsiBackbones}
\end{table}\vspace{-5pt}

\begin{table*}
    \centering
\caption{PathDino's performance, assessed for patch-level search accuracy and MV@5 macro average $F$-$1$ score, compared to various feature extractors. The lower-right section (grey values) indicates datasets that have been partially or fully included in the pretraining dataset TCGA.} \vspace{-5pt}
    \begin{adjustbox}{width=1.0\textwidth}
    \small
    \begin{tabular}{ccl|cccccccc|cccccccc|cccccc}
    \hline
    &&& \multicolumn{8}{c|}{Internal Datasets} & \multicolumn{14}{c}{Public Datasets}     \\\cline{4-25}  
    &&& \multicolumn{2}{c}{Private-Breast} &\multicolumn{2}{c}{Private-Liver}  &\multicolumn{2}{c}{Private-Skin} &\multicolumn{2}{c|}{Private-CRC} &\multicolumn{2}{c}{PANDA \cite{DataPANDA}} &\multicolumn{2}{c}{CAMELYON16 \cite{DataCAMELYON16}} &\multicolumn{2}{c}{BRACS \cite{DataBRACS}}&\multicolumn{2}{c}{DigestPath \cite{DataDigestPath}} &\multicolumn{2}{c|}{Kather \cite{DatasetKather}} &\multicolumn{2}{c}{PanNuke \cite{DataPanNuke}} &\multicolumn{2}{c}{WSSS4LUAD \cite{DataWSSS4LUAD}}\\
    \hline
    &&& Acc & MAF1 & Acc & MAF1 & Acc & MAF1 & Acc & MAF1 & Acc & MAF1 & Acc & MAF1 & Acc & MAF1 & Acc & MAF1 & Acc & MAF1 & Acc & MAF1 & Acc & MAF1 \\
    \hline
    \multirow{11}{*}{\rotatebox[origin=c]{90}{Pretrained on Natural Data}} & \multirow{6}{*}{\rotatebox[origin=c]{90}{CNN-based}}
                      &ResNet50 \cite{Resnet}& 32.5&	19.0	&63.8	&42.8&	68.9&	53.0	&47.1	&47.3	&31.0	&26.0 &62.5	&56.4&	47.8&	40.6&  86.7	&82.0	&97.5	&97	&74.7&	59&	75.2&	45.2    \\ 
                &&DenseNet121 \cite{DenseNet}& 31.4	&19.2&	65.0	&43.6&	68.7&	53.5	&47.6	&47.6&31.6	&26.7&62.6	&57.3	&49&	42.6&  88.7&	86	&98.5	&98.1&	77.8&	63.1&	77.7&	48.5   \\ 
    &&EfficientNet-b3-288 \cite{Efficientnet}& 29.7	&16.6&	64.5	&46.7&	67.4	&51.8&	46.5	&46.7	&30.8&	25.8 &61.5	&55.4&	49&	42&  89.8	&87.5&	96.1&	95.5&	70.8	&53.3	&78.0&	48.1    \\  
        &&EfficientNet-b5 \cite{Efficientnet}& 38.2&	25.6&	68.1	&48.6&	71.9&	55.9&	50.6	&51.0	&34.7	&29.4 &62.4	&56.1&	50.6&	43.9& 92.2&	91	&98.6&	98.3&	79.8&	64.5	&79.7&	49.1    \\ 
             &&ConvNext-b-224 \cite{Convnext} & 39.7	&28	&68.3&	50.1	&72.6	&58.2&	48.7	&48.8	&33.2&	28.2 &64.5	&59.8&	49.9	&43.4& 92.2&	90.7	&99.2&	99.1&	85.6&	73.9	&83.2	&52.3   \\ 
            &&ConvNext-xlarge \cite{Convnext}& 42.8&	28.7&	70.0	&51.8&	74.7&	60.3&	51.3&	51.6&34.2&	29.1 &	63.4&	57.7	&51.3&	44.6 & 93.2	&92	&99.5&	99.5&	90.4&	81.6	&84.2	&53.5   \\  \cline{2-25}
     & \multirow{5}{*}{\rotatebox[origin=c]{90}{Transformer}}
                     &ViT-b16-224 \cite{ViT}& 29.6	&16.7&	67.7&	49.2	&71.7&	55.5&	46.8	&46.9&	32.1	&26.8&62.0&	55.8	&49.1&	42.2 &89.3	&80.7&	98.2	&97.8	&79.4&	67	&70.4	&51.3    \\ 
              &&DinoV1-ViT-s16 \cite{DinoV1}& 36.6	&25.0	&70.3&	49.6&	71.4&	56.6&	49.9&	50.2&34.1	&29.7 &	63.2	&57.3	&51.3	&44.6& 92	&90.1&	99.5&	99.4&	89.8	&81.3&	83.9	&53.4    \\ 
              &&DinoV1-ViT-b16 \cite{DinoV1}& 38.1	&27.2	&71.3&	52.1	&72.2&	57.8	&50.8&	51&	34.7&	30.1&64.5&	58.4&	51.3&	44.4 &   91.9&	90&	99.7&	99.6	&91.5&	83	&84.7&	54.2 \\ 
              &&DinoV2-ViT-b14 \cite{DinoV2}& 31.8	&20.9&	68.4	&48.7	&69.8&	54.4	&48.1&	48.3	&31.4&	26.4&	60.2&53.3&	50.0&	42.6& 89.8	&86.5	&98.6&	98.4&	76.6&	64.5&	76.1&	66.5     \\ 
               &&CLIP - ViT-B/16 \cite{CLIP}&36.4	&26.8&	69.4&	49.8	&72.7&	57.7&	52.3	&52.6	&35.8&	31.0 &62.8	&56.7	&52.5	&45.5 & 90.0	&87.8	&98.4&	98.2	&79.1&	63.7&	79.2&	48.6    \\ \hline

    \multirow{12}{*}{\rotatebox[origin=c]{90}{Pret. On Histopathology Data}} & \multirow{6}{*}{\rotatebox[origin=c]{90}{CNN-based}}
     & Barlow-Twins-ResNet50  \cite{SSLPath}  & 50.8	&37.5&	76.0	&55.5&	72.2&	56.7	&56.1	&56.9&	46.0&	43.5	&63.9&	58.0	&54.8	&47.1&  95.2	&94.4&	\textcolor{gray}{99.7}	&\textcolor{gray}{99.6}&	\textcolor{gray}{91.8}&	\textcolor{gray}{85.3}	&\textcolor{gray}{86.2}	&\textcolor{gray}{54.9 }   \\ 
            && SwAV-ResNet50  \cite{SSLPath}  & 50.2&	37.5	&77.4	&60.1	&74.2&	59.6	&56.2	&56.9&	45.0&	42.1&	68.6&	63.2	&55.8	&48.4 & 95.3	&94.7&	\textcolor{gray}{99.6}	&\textcolor{gray}{99.5}	&\textcolor{gray}{90.6}	&\textcolor{gray}{82.5}&	\textcolor{gray}{82.8}&	\textcolor{gray}{51.5 }   \\ 
          & & MoCoV2-ResNet50 \cite{SSLPath}  & 51.9&	37.5&	76.7	&57.9&	72.9&	56.3&	54.6	&55.3	&	45.2&	42.3&65.0	&58.9	&54.6	&47.4 & 94.7&	94.0	&\textcolor{gray}{99.7}&	\textcolor{gray}{99.6}	&\textcolor{gray}{90.8}	&\textcolor{gray}{83.7}&	\textcolor{gray}{84.6}	&\textcolor{gray}{53.7 }   \\ 
       & & MuDiPath-ResNet50 \cite{MuDiPath}  &32.5&	20.9&	68.0	&47.2	&71.5&	55.6&	47.0&	47.2&	31.8	&27.0&	62.1&	57.0&	49.0	&42.0 & 89.4	&87.7&	\textcolor{gray}{98.9}&	\textcolor{gray}{98.5}	&\textcolor{gray}{80.6}&	\textcolor{gray}{68.9}	&\textcolor{gray}{81.0}	&\textcolor{gray}{50.7}    \\  
    && MuDiPath-DenseNet-101 \cite{MuDiPath}  & 36.6&	25.9	&69&	47.5&	72.0	&56.2&	49.4&	49.8&	33.3&	28.8&62.3&	56.4&	50.5&	43.5 & 91.6	&89.3	&\textcolor{gray}{99.4}	&\textcolor{gray}{99.2}&	\textcolor{gray}{88.8}&	\textcolor{gray}{79.7}&	\textcolor{gray}{82.8}&	\textcolor{gray}{52.5}    \\ 
                  &&KimiaNet \cite{kimianet}  & 46.8&	37.2	&78.2	&61.2&	76.3&	61.6&	56.0	&56.7	&45.1	 &  42.4  &71.9&	67.7&	56.8	&50.6& 95.0&	94.2	&\textcolor{gray}{99.4}	&\textcolor{gray}{99.3}&	\textcolor{gray}{94.3}&	\textcolor{gray}{88.6}&	\textcolor{gray}{82.4}&	\textcolor{gray}{51.4 }   \\ \cline{2-25}
    
& \multirow{5}{*}{\rotatebox[origin=c]{90}{Transformers}}
   &BiomedCLIP \cite{zhang2023large}& 34.1&	22.7	&67.8&	49.7&	72.1	&56.3	&47.6	&47.7&	32.5&	27.4&	61.3	&55.4	&50.6&	43.6 & 92.8	&91.3&	\textcolor{gray}{98.6	}&\textcolor{gray}{98.3}	&\textcolor{gray}{79.8}	&\textcolor{gray}{66.8}&	\textcolor{gray}{84.1}&	\textcolor{gray}{53.6 }  \\ 
          &&HIPT-ViT-s16 \cite{HIPT}& 37.8&	25.0&	70.6	&50.3&	71.5	&56.3	&49.2	&49.4&	33.8	&28.9&	67.2&	62	&50.1&	43.2& 89.3	&87.5	&\textcolor{gray}{98.7}&\textcolor{gray}{	98.3}&	\textcolor{gray}{88.6}&	\textcolor{gray}{78.2}&	\textcolor{gray}{81.0}&	\textcolor{gray}{50.5  }  \\ 
                  &&PLIP \cite{PLIP}& 44.1&	34.9&	72.0	&54.1&	75.2&	61.6&	\textbf{57.8}&	\textbf{58.4}&	43.0&	39.3&68.8&	62.9&	55.4&	48.2& 94.7	&93.7	&\textcolor{gray}{97.2}	&\textcolor{gray}{97.0}	&\textcolor{gray}{82.3}&	\textcolor{gray}{68.6}&	\textcolor{gray}{78.2}&	\textcolor{gray}{48.5 }   \\ 
           &&iBOT-Path  \cite{iBOT}& 50.2	&42.1	&78.0	&65.2	&76.8&	62.4	&55.9&	56.5&	41.6&	37.9&	69.9&	64.4&	57.8&	51.2& 95.2&	94.3	&\textcolor{gray}{99.9}	&\textcolor{gray}{99.9	}&\textcolor{gray}{97.7}&\textcolor{gray}{	93.6}	&\textcolor{gray}{87.1}&	\textcolor{gray}{55.7 }   \\ 
 &&DinoSSLPathology-8 \cite{SSLPath} & 47.1	&36.3&	77.0&	59.7&76.1&	61.4&	56.0	&56.6&	39.8	&35.3	&67.8	&60.8	&56.0&	49.0 & 95.7	&95.2	&\textcolor{gray}{99.9}&\textcolor{gray}{	99.9}&	\textcolor{gray}{96.6}	&\textcolor{gray}{92.2}&	\textcolor{gray}{88.1}	&\textcolor{gray}{56.7 }  \\ \cline{3-25} \rowcolor{LightBlue}\hline
                &&PathDino-224 (ours) & 44.5&	38.7&	77.2	&61.6	&76.0&	61.4&	52.7&	53.2&	40.1&	36.0&	71.6&	66.9	&55.1	&48.6& 95.8&	95.0	&\textcolor{gray}{99.9}&	\textcolor{gray}{99.8}&	\textcolor{gray}{96.3	}&\textcolor{gray}{90.7}&	\textcolor{gray}{86.9	}&\textcolor{gray}{55.7  } \\ \rowcolor{LightBlue}
              &&PathDino-512 (ours)  & \textbf{55.1}	&\textbf{49.1}	&\textbf{82.7}	&\textbf{69.5}&	\textbf{77.2}&	\textbf{63.6}&	57.4	&58.1&	\textbf{48.3}	&\textbf{46.3}	&\textbf{75.1}&	\textbf{70.4}	&\textbf{59.3}&	\textbf{52.6} & \textbf{96.8}	&\textbf{96.2}	&\textcolor{gray}{99.9}&	\textcolor{gray}{99.9}	&\textcolor{gray}{96.6	}&\textcolor{gray}{91.1}&	\textcolor{gray}{86.7}&	\textcolor{gray}{55.4 }  \\  \hline

    \end{tabular}
    \end{adjustbox}
    \label{tab:extendedResults_MV5}
\end{table*}

\begin{figure*}
    \centering 
    \includegraphics[width=.95\textwidth]{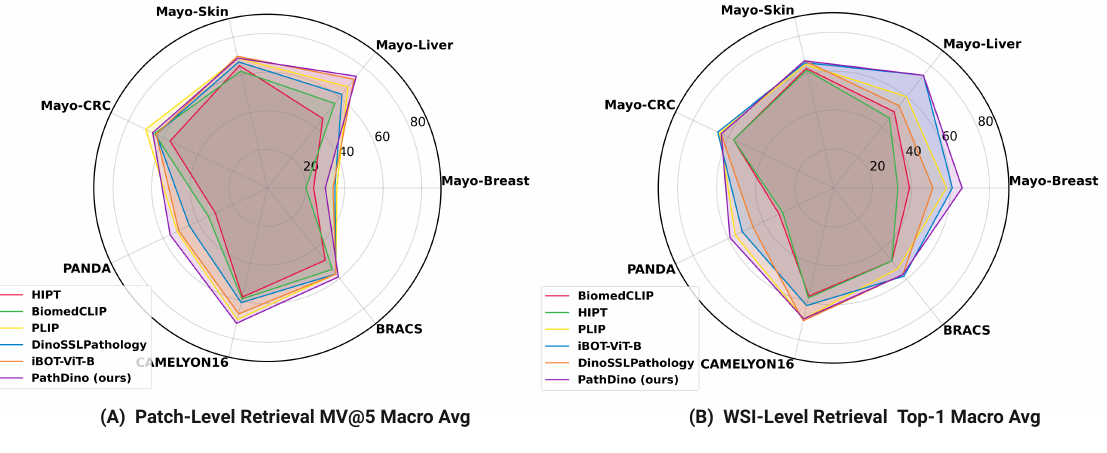}   \vspace{-15pt}
    \caption{Performance of selected Transformer-based histopathological feature extractors including HIPT, BiomedCLIP, PLIP, DinoSSLPath, iBOT, and PathDino. The performance is represented as the macro average of the $F1$ score for the MV@5: (A) the performance of patch-level retrieval, (B)  the performance of WSI-level retrieval. }  
    \label{fig:transformersPerformance}
\end{figure*}\vspace{-3pt}

\begin{table*}
    \centering
\caption{5-Fold Cross-Validation: Macro-F1 in Histopathology. Right side: TCGA-related datasets (see the Supplementary File).} \vspace{-7pt}
    \begin{adjustbox}{width=1.0\textwidth}
    \small
    \begin{tabular}{ccl|cccc|cccc|cccc}
    \hline
    &&& \multicolumn{4}{c|}{Internal Datasets} & \multicolumn{8}{c}{Public Datasets}     \\\cline{4-15} 
    &&& Private-Breast & Private-Liver &  Private-Skin & Private-CRC & PANDA \cite{DataPANDA} & CAMELYON16 \cite{DataCAMELYON16}& BRACS \cite{DataBRACS}& DigestPath \cite{DataDigestPath}& Kather \cite{DatasetKather}& PanNuke \cite{DataPanNuke}& WSSS4LUAD \cite{DataWSSS4LUAD}\\
    \hline
& \multirow{5}{*}{\rotatebox[origin=c]{90}{Transformers}}&BiomedCLIP \cite{zhang2023large}
& 38.82$\pm$1.64&	48.44$\pm$1.15&	56.62$\pm$0.61&	55.89$\pm$2.57& 25.97$\pm$0.34 &		58.19$\pm$4.99&	41.89$\pm$0.93& 92.07$\pm$2.33	&\textcolor{gray}{94.89$\pm$0.84}&\textcolor{gray}{	37.81$\pm$2.12}&\textcolor{gray}{	69.98$\pm$8.12}&\\ 
&&HIPT-ViT-s16 \cite{HIPT} &43.08$\pm$6.27&	59.31$\pm$5.08&	59.47$\pm$2.85&	48.69$\pm$4.62&	25.65$\pm$1.38&	61.56$\pm$4.25&	42.02$\pm$8.82& 84.66$\pm$7.17&\textcolor{gray}{	96.81$\pm$0.69	}&\textcolor{gray}{42.17$\pm$3.80	}&\textcolor{gray}{64.26$\pm$7.50}\\ 
&&PLIP \cite{PLIP} &46.07$\pm$3.20&	50.78$\pm$1.48&	62.48$\pm$1.11&	64.11$\pm$2.70&31.53$\pm$0.47&69.67$\pm$1.45&	46.72$\pm$1.05& 92.07$\pm$2.91&\textcolor{gray}{	90.90$\pm$1.63}&\textcolor{gray}{	27.77$\pm$2.54	}&\textcolor{gray}{61.51$\pm$7.19}\\ 
&&iBOT-Path  \cite{iBOT} &85.12$\pm$1.74&	84.37$\pm$1.31	&\textbf{73.09$\pm$0.39}	&68.38 $\pm$0.52	&\textbf{32.95$\pm$0.86}&	73.76$\pm$1.67&	\textbf{56.52 $\pm$1.96}& 95.67$\pm$2.17	&\textcolor{gray}{99.81$\pm$0.17	}&\textcolor{gray}{95.76$\pm$1.78	}&\textcolor{gray}{73.31$\pm$5.91}\\ 
&&DinoSSLPathology-8 \cite{SSLPath} &77.59$\pm$2.17&	74.25$\pm$4.56&	66.98$\pm$1.00&	58.17$\pm$4.77&28.75$\pm$2.31&70.61$\pm$1.81&	46.42$\pm$5.59& 94.50$\pm$1.91&\textcolor{gray}{	99.68$\pm$0.11}&\textcolor{gray}{	86.17$\pm$2.61}&\textcolor{gray}{	76.30$\pm$9.60}\\  \cline{3-15} \rowcolor{LightBlue}\hline
&&PathDino-224 (ours)  & 78.06$\pm$4.03&	74.34$\pm$4.98&	64.89$\pm$2.14&	60.65$\pm$2.23&	27.74$\pm$2.44&	69.26$\pm$4.94&	46.58$\pm$3.78&94.03$\pm$3.06&\textcolor{gray}{	99.66$\pm$0.19}&\textcolor{gray}{	81.03$\pm$2.51	}&\textcolor{gray}{74.47$\pm$9.05}\\ \rowcolor{LightBlue}
&&PathDino-512 (ours)  &\textbf{88.57$\pm$3.08}&	\textbf{86.35$\pm$5.33}&	71.36$\pm$1.64&\textbf{	70.47$\pm$2.47}&	32.08$\pm$2.57&	\textbf{79.61$\pm$1.00}&	52.59$\pm$3.21& \textbf{95.82$\pm$2.26}&\textcolor{gray}{	99.65$\pm$0.11	}&\textcolor{gray}{84.79$\pm$3.14}	&\textcolor{gray}{72.69$\pm$7.60}\\ \hline

    \end{tabular}
    \end{adjustbox}
    \label{tab:patchClassificationMacro}
\end{table*}

\subsection{PathDino - WSI-Level Search}
\label{sec:wsi}
Table \ref{tab:wsiBackbones} highlights the performance of several feature extractors across various private and public datasets using the proposed FPS patching method and the \emph{median of minimum} Euclidean distances proposed in Yottixel \cite{yottixel}. Across both private and public datasets, PathDino-512 demonstrates competitive to superior performance. PathDino-512 achieves an exceptional $83\%$ top-1 accuracy in the dataset Private-Liver, outperforming other models like HIPT, and iBOT-Path (student), which attain $68\%$ and $79\%$, respectively. Even in a difficult case like Private-Skin, PathDino-512 reaches a $78\%$ top-1 accuracy, competing with DinoSSLPathology which provides $78\%$. Notably, in the public dataset PANDA, PathDino-512 achieves a $58\%$ top-1 accuracy, significantly outperforming both CNN-based and Transformer-based models like HIPT which only reach $32\%$. The macro average $F1$ score also consistently favors PathDino-512. These empirical findings prove PathDino-512 is a robust and highly efficient model for WSI-level retrieval. More results for the macro average $F1$ score of Top1, MV@3, MV@5, along with accuracy of MV@3, and MV@5 are reported in Suppl-Tables \ref{tab:WSIMacroTop1}, \ref{tab:WSIMacroMV3}, \ref{tab:WSIMacroMV5}, \ref{tab:WSIAccMV3}, and \ref{tab:WSIAccMV5}, respectively.

\subsection{PathDino - Patch-Level Search}
\label{sec:patchSearch}
The results presented in Table \ref{tab:extendedResults_MV5} provide an extensive comparative analysis of models in patch-level histopathology image search. The standout performer is our proposed model, PathDino-512. The model not only outperforms others in terms of accuracy but also establishes new benchmarks in the macro average $F1$ score, a critical metric for robust evaluation.
For private datasets such as Private-Breast and Private-Liver, PathDino-512 achieves the highest accuracy rates of $55.1\%$ and $82.7\%$, respectively. More remarkably, it tops the macro average $F1$ score with $49.1\%$ and $69.5\%$ in the same datasets. These findings extend to public datasets like PANDA and CAMELYON16, where PathDino-512 records accuracy and macro average $F1$ scores of $48.3\%$ and $46.3\%$, and $75.1\%$ and $70.4\%$, respectively.

While it is important to note the strong performance of models like iBOT-Path and DinoSSLPathology, especially for public datasets, PathDino-512 consistently outperforms them across multiple metrics and datasets. We analyzed the patch embedding variance as shown in Fig. \ref{fig:FigEmbeddingVariance}. We compare PathDino against HIPT and DinoSSLPathology as they have the same embedding size (\ie, $384$). Notably, PathDino capitalizes on an expanded set of components within the feature vector to accurately represent the inferred histopathology patch. Fig. \ref{fig:FigAttenCompare3Models} visually compares their attention performance in  which PathDino shows better attentions.

\subsection{PathDino - Patch-level 5-Fold Cross-Validation}
\label{sec:patch5FoldVal}
In Table \ref{tab:patchClassificationMacro} detailing 5-fold cross-validation results, a thorough quantitative comparison of macro-averaged $F1$ scores is presented for an assortment of models across multiple private and public datasets. We only report the performance of histopathology Transformer-based models here. The detailed measurements of macro average $F1$ scores and accuracy values are available in Suppl-Tables \ref{tab:crossValidMacro} and \ref{tab:crossValidAcc}, respectively.

On the internal datasets like Private-Breast, Private-Liver, and Private-CRC, our proposed model, PathDino-512, achieves standout performance with $F1$ scores of 88.57±3.08, 86.35±5.33, and 70.47±2.47, respectively. These scores are markedly higher than the next best models, such as iBOT-Path, which reaches $F1$ scores of $85.12\pm1.74$ in Private-Breast and $84.37\pm1.31$ in Private-Liver. In the realm of public datasets, PathDino-512, and iBOT-Path show competitive results where PathDino leads with an $F1$ score of $79.61\pm1.00$ in CAMELYON16, outperforming iBOT-Path, which scores $73.76\pm1.67$ in the same dataset. Interestingly, iBOT-Path excels in Private-Skin with an $F1$ score of $73.09\pm0.39$, the highest among all models for that specific dataset.

\section{Conclusions}

This paper presented a new approach to WSI analysis, addressing two pivotal challenges that have long stymied advancements in this field—computational efficiency and diagnostic fidelity. We introduced a \emph{fast patch selection (FPS)} algorithm that reliably identifies a compact yet highly informative subset of patches, thereby significantly reducing computational overhead without compromising diagnostic inclusion. Additionally, we unveiled a new Transformer-based model structure for histopathological image analysis, \emph{PathDino}, that only contains 5 small transformer blocks. Finally, we presented a rotation-agnostic self-supervised learning, \emph{HistoRotate}, tailored for histopathological representation learning. Through training the proposed \emph{PathDino} using the proposed \emph{HistoRotate} and rigorously validating them with 12 diverse datasets, we showed that our lightweight transformer along with our training recipe effectively mitigates issues of overfitting that are prevalent in this domain. Our dual-pronged approach has demonstrated competitive to superior performance compared to the state-of-the-art methods. 

\noindent\textbf{Limitations}: In contrast to natural images, magnification plays an important role in histopathological images. Our training dataset only included patches in $20X$ magnification from TCGA. Thus, more tuning for multi-resolution training may provide better results.

\noindent\textbf{Broader Impacts}: The proposed methods for whole slide image analysis have the potential to improve the diagnosis and prognosis of various diseases by providing accurate and reliable information on tissue morphology and cellular characteristics. With the widespread use of digital pathology workflows in clinical practice, these methods can reduce the workload and human errors of pathologists. Furthermore, quantifying tissue morphologies through accurate and valid image analysis method help with reducing intra- and inter-observer variability within the medical field. The proposed methods can also contribute to the advancement of histopathological image analysis by providing robust image representations.

\noindent\textbf{Acknowledgments}:  
 The authors thank 
Mark Zarella,
Wenchao Han,
Sobhan Hemati, 
Nneka Comfere, 
Dennis Murphree, 
Chady Meroueh, 
Saba Yasir, 
Aaron Mangold, 
Lisa Boardman, 
Vijay H. Shah, and 
Joaquin J. Garcia 
for their valuable insights, discussions, and suggestions.

{\small
\bibliographystyle{unsrt}
\bibliography{egbib}
}

\clearpage
\onecolumn
\appendix
\beginsupplement

    \centering
    {\Large
    \textbf{Rotation-Agnostic Image Representation Learning for Digital Pathology}}
    \vspace{1cm}
    
    {\Large
    \textit{(Supplementary File)}}
    \vspace{1cm}


{\large \textbf{Figures}}

\begin{figure*}[!h] 
    \centering 
    \includegraphics[width=.86\textwidth]{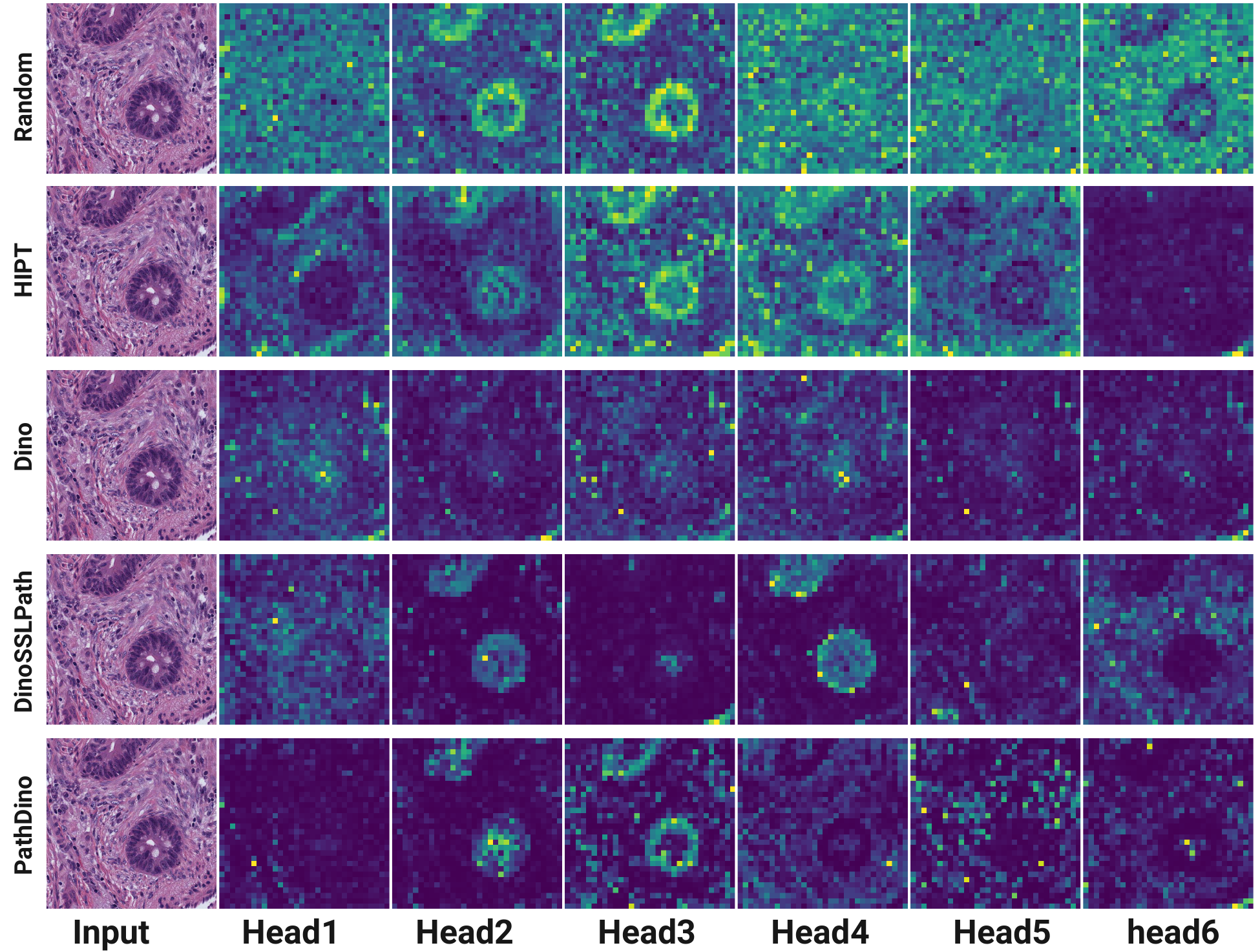}  
    \caption{Attention visualization of the proposed PathDino as compared to other ViT-samll models.}  
    \label{fig:pathDinoAttentionSmalls}
\end{figure*}

\begin{figure*}[!h] 
    \centering 
    \includegraphics[width=.86\textwidth]{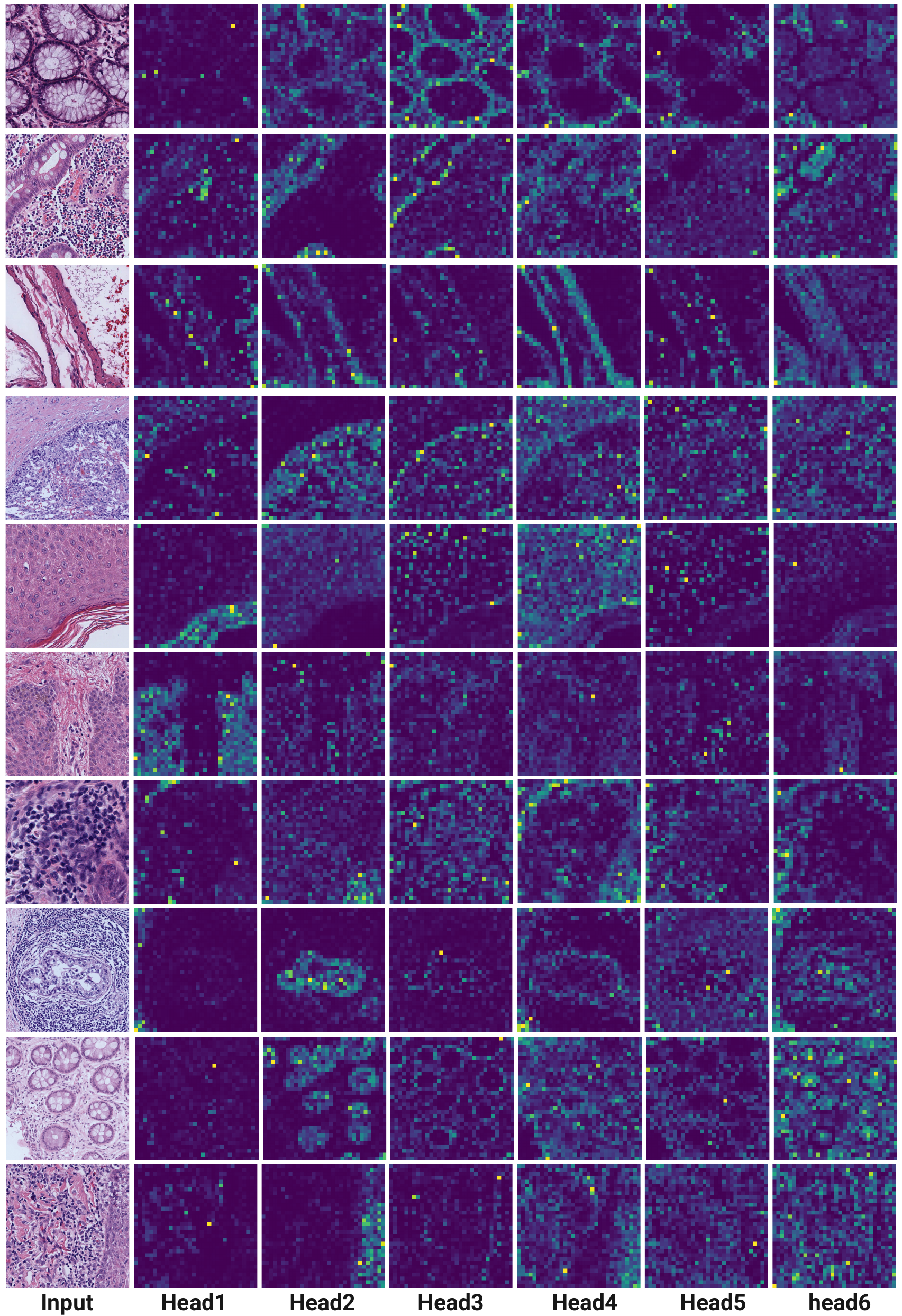}  
    \caption{Attention visualization of the proposed PathDino transformer heads.}  
    \label{fig:pathDinoAttentionHeads}
\end{figure*}

\begin{figure*}[!h] 
    \centering 
    \includegraphics[width=.95\textwidth]{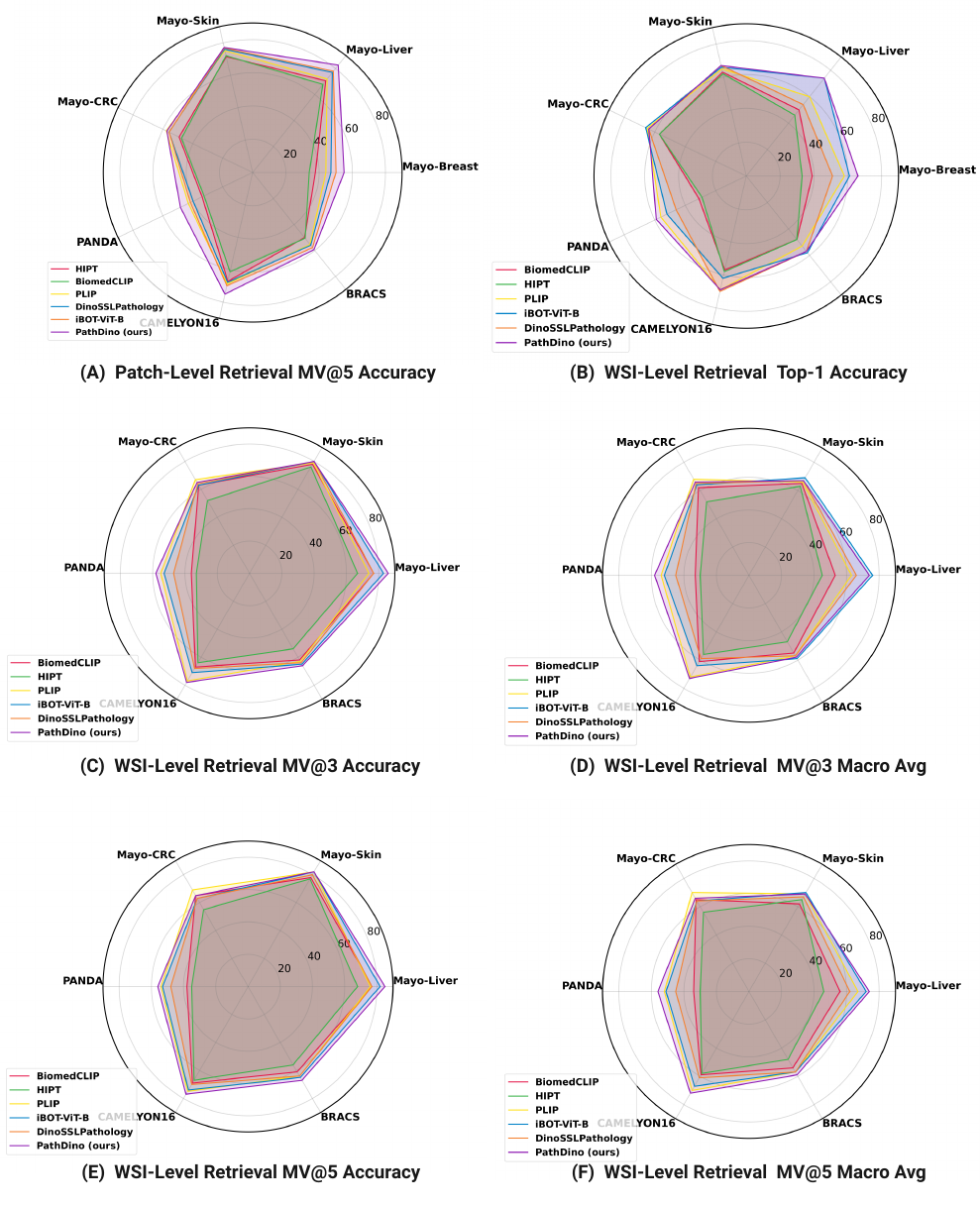}  
    \caption{Performance of selected Transformer-based histopathology feature extractors including HIPT, BiomedCLIP, PLIP, DinoSSLPath, iBOT, and PathDino. The performance is represented as the macro average of the $F_1$ score for the MV@5 in the patch-level retrieval settings.   }  
    \label{fig:transformersPerformanceSuppl}
\end{figure*}

\begin{figure*}[!h] 
    \centering 
    \includegraphics[width=.95\textwidth]{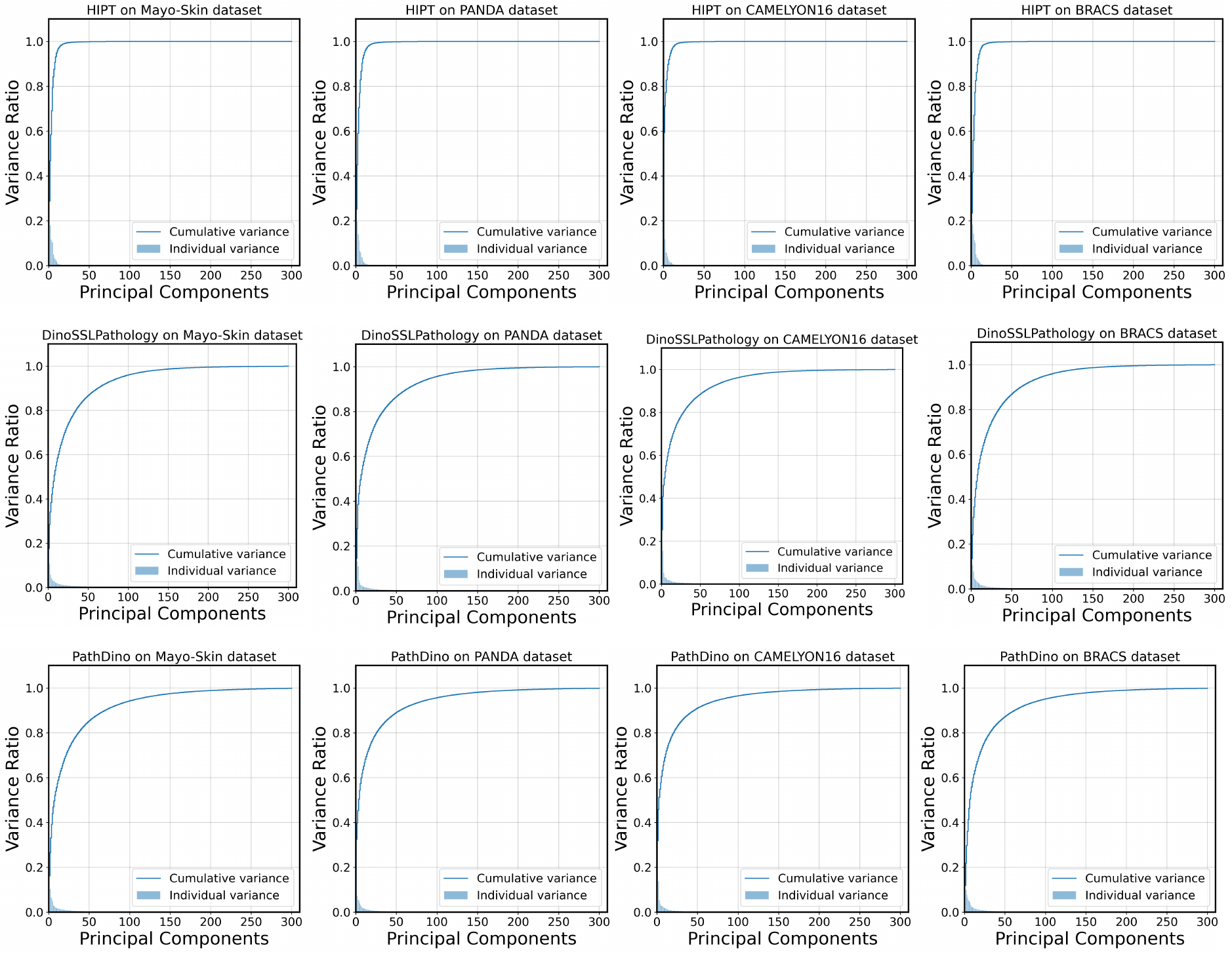}  
    \caption{Embedding variance analysis of three selected Transformer-based histopathology feature extractors which have output vector size of $384$ including HIPT, DinoSSLPath, and our PathDino.   }  
    \label{fig:embeddingVariance}
\end{figure*}

\begin{enumerate}
\item Fig. \hyperref[Attention visualization]{\ref{fig:pathDinoAttentionSmalls}}: Attention Visualization - PathDino vs. Its Counterparts. 
\item Fig. \hyperref[]{\ref{fig:pathDinoAttentionHeads}}: PathDino Attention Visualization.
\item Fig. \hyperref[]{\ref{fig:transformersPerformanceSuppl}}: Radar diagram - PathDino Performance vs. Its Counterparts. 
\item Fig. \hyperref[]{\ref{fig:embeddingVariance}}: Embedding Variance Analysis -  PathDino, HIPT, DinoSSLPath.
\end{enumerate}
\vspace{12pt}

{\large \textbf{Tables}}

\begin{enumerate}
\item Table \hyperref[]{\ref{tab:model-attributes}}: Backbones Attributes.

\item Table \hyperref[]{\ref{tab:wsiLevelFPSvsYottixelAcc}}: FPS Vs. Yottixel WSI-level Accuracy.

\item Table \hyperref[]{\ref{tab:wsiLevelFPSvsYottixelF1}}: FPS Vs. Yottixel WSI-level Macro Average F1-Score.

\item Table \hyperref[]{\ref{tab:crossValidAcc}}: PathDino Vs. Counterparts F-fold Cross-Validation (Accuracy).

\item Table \hyperref[]{\ref{tab:crossValidMacro}}: PathDino Vs. Counterparts F-fold Cross-Validation (Macro Average F1-Score).

\item Table \hyperref[]{\ref{tab:privateDataset}}: Private Datasets Properties.

\item Table \hyperref[]{\ref{tab:publicDataset}}: Public Datasets Properties.

\item Table \hyperref[]{\ref{tab:WSIAccTop1}}: PathDino Vs. Counterparts - WSI-level - Top-1 - Accuracy.

\item Table \hyperref[]{\ref{tab:WSIMacroTop1}}: PathDino Vs. Counterparts - WSI-level - Top-1 - Macro Average F1-Score.

\item Table \hyperref[]{\ref{tab:WSIAccMV3}}: PathDino Vs. Counterparts - WSI-level - MA@3 - Accuracy.

\item Table \hyperref[]{\ref{tab:WSIMacroMV3}}: PathDino Vs. Counterparts - WSI-level - MA@3 -  Macro Average F1-Score.

\item Table \hyperref[]{\ref{tab:WSIAccMV5}}: PathDino Vs. Counterparts - WSI-level - MA@5 - Accuracy.

\item Table \hyperref[]{\ref{tab:WSIMacroMV5}}: PathDino Vs. Counterparts - WSI-level - MA@5 -  Macro Average F1-Score.

\end{enumerate}

\vspace{12pt}

{\large \textbf{Algorithms}}
\begin{enumerate}
\item Algorithm \hyperref[]{\ref{alg:rotation-agnostic-pipeline}}: HistoRotate: $360^\circ$ Rotation Augmentation Method.
\end{enumerate}

\twocolumn

\begin{table*}[!h]
    \centering
    \caption{Summary of Model Attributes. It is worth noting that DinoV2, CLIP, and DinoSSLPath also trained with CNN-based backbones such as ResNet50, however, we only employ the Transformer-based backbones in our comparisons. Floating Point Operations Per Second (FLOPs) are used to quantify the computational complexity of models. Note that we specified the data size used for the exact pre-trained models in our study rather than the general mentioned in the corresponding papers. For example, DinoSSLPath \cite{SSLPath} was trained on $32.6$M patches ($19$M from TCGA and $13.6$M from a TULIP, which is a private dataset), however, the publically available pre-trained models are trained only on TCGA. Thus, we report $19$M for pretraining DinoSSLPath, SwAV-ResNet50, MoCoV2-ResNet50, and Barlow-Twins-ResNet50. PubMed consists of a total of $15,282,336$, however, the training set is $13.9$M.}
    \begin{adjustbox}{width=1.0\textwidth}
    \small
    \label{tab:model-attributes}
    \begin{tabular}{llcccrcccrrr}
        \toprule\hline
        &Model & Pretrained On & Pretraining Domain & Modality & Training Data Size  & Learning Paradigm & Input Dim. & Output Embedding& Model Size & FLOPs  \\
        \midrule 
        \multirow{11}{*}{\rotatebox[origin=c]{90}{\large CNN}} 
        & ResNet50 - \cite{Resnet}& ImageNet-1k & Natural Images & Image & $1,200,000$ & Supervised & $224\times224$ &2048&  $23,527,264$ & $4,374,897,664$ \\
        &DenseNet121 - \cite{DenseNet}& ImageNet-1k & Natural Images & Image & $1,200,000$&  Supervised  & $224\times224$ &$1024$& $6,870,208$ & $2,833,364,480$  \\
        &EfficientNet-b3-288 - \cite{Efficientnet}& ImageNet-1k & Natural Images & Image & $1,200,000$&  Supervised  & $288\times288$ &$1536$& $10,608,936$& $1,587,788,048$ \\
        &EfficientNet-b5 - \cite{Efficientnet}& ImageNet-1k & Natural Images & Image &  $1,200,000$  & Supervised &$448\times448$  &$2048$&  $28,168,048$ & $9,402,729,536$  \\
       & ConvNext-B \cite{Convnext}& ImageNet-21k & Natural Images & Image &$14,000,000$  & Supervised   & $224\times224$ & $1024$&$87,510,272$ & $15,353,709,568$ \\
        &ConvNext-xlarge - \cite{Convnext}& ImageNet-21k & Natural Images & Image &  $14,000,000$  & Supervised & $224\times224$ &$2048$& $348,035,584$ & $ 60,918,990,848$ \\
        & SwAV-ResNet50 - \cite{SSLPath}& TCGA & Histopathology Images & Image & $19,000,000$  &  Self-supervised   & $1024\times768$ &$2048$&  $23,508,032$ &  $64,757,958,656$ \\
        &MoCoV2-ResNet50 - \cite{SSLPath}& TCGA & Histopathology Images & Image & $19,000,000$ &   Self-supervised  & $1024\times768$  &$2048$& $23,508,032$ & $64,757,958,656$ \\
        &MuDiPath-ResNet50 - \cite{MuDiPath}& TCGA & Histopathology Images & Image &  $882,800$  & Supervised  & $224\times224$ &$2048$&$25,557,032$  & $4,131,592,192$  \\
        &MuDiPath-DenseNet-101 - \cite{MuDiPath}& TCGA & Histopathology Images & Image &  $882,800$  & Supervised & $224\times224$ &$1024$&$6,953,856 $ & $2,895,983,104$ \\
        &KimiaNet - \cite{kimianet}& TCGA & Histopathology Images & Image & $240,000$  & Supervised & $1000 \times 1000$ & $1024$&$6,953,856$ & $ 57,471,584,640$ \\
        &Barlow-Twins-ResNet50 - \cite{SSLPath}& TCGA & Histopathology Images & Image & $19,000,000$ &   Self-supervised  & $1024\times768$  &$2048$& $23,508,032$ &$ 64,757,958,656$ \\\hline
        \multirow{11}{*}{\rotatebox[origin=c]{90}{\large Transformers}}         &ViT-B16 \cite{ViT}& ImageNet-1K & Natural Images & Image  &  $1,200,000$ & Supervised & $224\times224$ &$768$& $85,646,592$ & $16,862,862,336$ \\
       & DinoV1-ViT-s16 - \cite{DinoV1}&  ImageNet-1k & Natural Images & Image & $1,200,000$  &  Self-supervised & $224\times224$ &$384$ & $21,589,632 $& $ 4,248,399,360$ \\
        &DinoV1-ViT-b16 - \cite{DinoV1}& ImageNet-1k & Natural Images & Image &  $1,200,000$  &  Self-supervised &  $224\times224$& $768$& $85,646,592$ & $16,862,862,336 $ \\
        &DinoV2-ViT-b14 - \cite{DinoV2}& Internet & Natural Images & Image & $142,000,000$   & Self-Supervised &  $224 \times 224$ & $768$&$85,508,352$ &$ 21,963,549,696$ \\
       & CLIP - ViT-B/16 - \cite{CLIP}& Internet & Natural Image-Text & Image-Text & $400,000,000$  & Contrastive Learning & $224 \times 224$ &$512$ &$ 85,646,592$ & $16,862,862,336$\\
        &BiomedCLIP - \cite{zhang2023large}& PMC-15M & Medical (PubMed) & Image-Text & $13,900,00$   & Contrastive Learning & $224 \times 224$ & $512$& $85,646,592$ & $16,862,862,336$\\
        &HIPT-ViT-s16 \cite{HIPT}& TCGA & Histopathology Images & Image & $104,000,000$  & Self-supervised& $256\times256$ &$384$&$21,589,632$  &$5,542,417,920 $ \\
        &PLIP \cite{PLIP}& OpenPath & Histopathology (Twitter) & Image-Text & $208,414$  & Contrastive Learning & $224 \times 224$ &$512$&  $85,646,592$ & $16,862,862,336$ \\
        &iBOT-Path  \cite{iBOT}& TCGA & Histopathology Images & Image & $40,000,000$ & Self-supervised& $224\times224$ & $768$              &$85,646,592$  & $16,862,862,336$ \\
        &DinoSSLPathology-8 \cite{SSLPath}& TCGA & Histopathology Images & Image & $19,000,000$  & Self-supervised & $224 \times 224$ & $384$&$21,368,448$  & $16,756,372,992$ \\
        \rowcolor{Gray}&PathDino-224 (ours)& TCGA & Histopathology Images & Image & $2,118,068$ &Self-supervised & $224\times224$ &$384$ & $9,168,384$     & $1,804,061,184 $ \\
        \rowcolor{Gray}&PathDino-512 (ours)& TCGA & Histopathology Images & Image & $6,087,558$  &Self-supervised &$512\times512$  &$384$&$9,168,384 $ &$9,387,852,288$  \\
        \hline\bottomrule
    \end{tabular}
    \end{adjustbox}
\end{table*}

\begin{table*}[htbp]
    \centering
\caption{(accuracy) Patient matching outcomes across four internal and three public datasets for classification, subtyping, and grading tasks, utilizing the Yottixel framework and leave-one-patient-out method.}
    \begin{adjustbox}{width=1.0\textwidth}
    \small
    \begin{tabular}{clcc>{\columncolor{Gray}}c|c>{\columncolor{Gray}}c|c>{\columncolor{Gray}}c|c>{\columncolor{Gray}}c|c>{\columncolor{Gray}}c|c>{\columncolor{Gray}}c|c>{\columncolor{Gray}}c} 
    \hline\hline
    & \multirow{2}{*}{Dataset}&&\multicolumn{2}{c}{DinoV2 \cite{DinoV2}} & \multicolumn{2}{c}{CLIP \cite{CLIP}} & \multicolumn{2}{c}{BiomedCLIP \cite{zhang2023large}} & \multicolumn{2}{c|}{PLIP \cite{PLIP}} & \multicolumn{2}{c}{KimiaNet \cite{kimianet}} & \multicolumn{2}{c}{DinoSSLPath \cite{SSLPath}} & \multicolumn{2}{c}{PathDino} \\
    \cline{4-17}
&&& Yottixel & FPS & Yottixel & FPS & Yottixel & FPS & Yottixel & FPS & Yottixel & FPS & Yottixel & FPS & Yottixel & FPS\\
\hline\hline
\multirow{10}{*}{\rotatebox[origin=c]{90}{\textbf{Internal Data}}} & \multirow{1}{*}{Private-Breast} & Top 1 &36&\textbf{53} & 45&\textbf{49} & 47&\textbf{47} & 55& \textbf{58}& \textbf{56}& 51& \textbf{59} &58&63& \textbf{68} \\\cline{2-17}

& \multirow{3}{*}{Private-Liver} & Top 1 &71 &\textbf{71} &63  &\textbf{67} & 70 &\textbf{74} & 70 & \textbf{73}& 76 &\textbf{78} & \textbf{75} &74&81& \textbf{83} \\
& & MV@3 &73 & \textbf{74}& 69 &\textbf{69} &75 &\textbf{77} & \textbf{76} &74 & 79 &\textbf{80} & 77 &\textbf{77}&86&\textbf{86} \\
& & MV@5 &70 &\textbf{72} & \textbf{72} &71 &74 & \textbf{77}& 73 & \textbf{76}& \textbf{80} & 78& \textbf{81} & 77&\textbf{87}&85\\\cline{2-17}

& \multirow{3}{*}{Private-Skin} & Top 1 &59 & \textbf{72}& 68 & \textbf{75}&68 &\textbf{75} & 72 & \textbf{75}& \textbf{78} & 75& \textbf{79} &78&\textbf{79}&78 \\
& & MV@3 &66 &\textbf{77} &73 & \textbf{79}&73 &\textbf{78} & 77 &\textbf{79} & 81 &\textbf{81} & 80 & \textbf{80}&\textbf{81}&80\\
& & MV@5 &68 &\textbf{87} & 77 &\textbf{80} &76 &\textbf{78} & 80 &\textbf{82} & 82 &\textbf{82} & 80 &\textbf{80} &81&\textbf{82}\\\cline{2-17}

& \multirow{3}{*}{Private-CRC} & Top 1 &52 & \textbf{56}&\textbf{57} & 56&55 & \textbf{58}& 60 & \textbf{64}& 60 & \textbf{64}& 60 &\textbf{63}&57& \textbf{63}\\
& & MV@3 &52 &\textbf{59} & \textbf{59} &58 &60 & \textbf{63}& 61 &\textbf{67} & 60 &\textbf{65} & 61 & \textbf{64}&60&\textbf{65}\\
& & MV@5 &55 & \textbf{59}& 56 &\textbf{60} &59 & \textbf{65}& 62 & \textbf{69}& 60 &\textbf{62} & 63 &\textbf{63}&61&\textbf{65} \\\hline\hline

\multirow{9}{*}{\rotatebox[origin=c]{90}{\textbf{Public Data}}} & \multirow{3}{*}{PANDA \cite{DataPANDA}} & Top 1 &\textbf{35} &31 &35  & \textbf{36}&33 &\textbf{34} & 53 & \textbf{56}& \textbf{58} & 57& \textbf{48} &47 &\textbf{59}&58\\
& & MV@3 &\textbf{33} &32 & 37 &\textbf{38} &36 &\textbf{36} & 53 & \textbf{55}& \textbf{58} &56 & \textbf{50} & 47&58&\textbf{58}\\
& & MV@5 &35 & \textbf{35}&39 & \textbf{40}&38 &\textbf{38} & 53 & \textbf{54}& \textbf{56} &54 & \textbf{51} &48&\textbf{58}&56 \\\cline{2-17}

& \multirow{3}{*}{CAMELYON16 \cite{DataCAMELYON16}} & Top 1 &57 & \textbf{61}&66 &\textbf{67} &60 &\textbf{61} & 70 &\textbf{73} & 75 & \textbf{76}& 62 &\textbf{74}&\textbf{76}& 73\\
& & MV@3 &58 & \textbf{58}& 66 &\textbf{67} &58 &\textbf{67} & 71 &\textbf{77} & 72 & \textbf{78}& \textbf{74} &68&77& \textbf{78}\\
& & MV@5 &57 & \textbf{63}& 64 &\textbf{69} &64 &\textbf{69} & 70 &\textbf{75} & 79 &\textbf{81} & \textbf{75} &70&\textbf{78}&77 \\\cline{2-17}

& \multirow{3}{*}{BRACS \cite{DataBRACS}} & Top 1 &\textbf{53} & 51&53 &\textbf{58} &\textbf{56} &55 & \textbf{62} &60 & \textbf{66} & 62& \textbf{62} &61 &\textbf{65}&64\\
& & MV@3 &\textbf{55} &53 & 58 &\textbf{60} &58 &\textbf{62} & \textbf{64} &63 & \textbf{66} & \textbf{64}& 61 &\textbf{64}&65&\textbf{66} \\
& & MV@5 &\textbf{58} &56 &59 &\textbf{61} &59 &\textbf{61} & \textbf{66} &64 & \textbf{67} &66 & 62 &\textbf{64} &66&\textbf{67}\\
    \hline\hline
    \end{tabular}
    \end{adjustbox}
    \label{tab:wsiLevelFPSvsYottixelAcc}
\end{table*}

\begin{table*}[htbp]
    \centering
\caption{(Macro Avg) Patient matching outcomes across four internal and three public datasets for retrieval, subtyping, and grading tasks, utilizing the FPS vs Yottixel framework and leave-one-patient-out method on Macro Average $F$-$1$ score.}
    \begin{adjustbox}{width=1.0\textwidth}
    \small
    \begin{tabular}{clcc>{\columncolor{Gray}}c|c>{\columncolor{Gray}}c|c>{\columncolor{Gray}}c|c>{\columncolor{Gray}}c|c>{\columncolor{Gray}}c|c>{\columncolor{Gray}}c|c>{\columncolor{Gray}}c}
    \hline\hline
    & \multirow{2}{*}{Dataset}&&\multicolumn{2}{c}{DinoV2 \cite{DinoV2}} & \multicolumn{2}{c}{CLIP \cite{CLIP}} & \multicolumn{2}{c}{BiomedCLIP \cite{zhang2023large}} & \multicolumn{2}{c|}{PLIP \cite{PLIP}} & \multicolumn{2}{c}{KimiaNet \cite{kimianet}} & \multicolumn{2}{c}{DinoSSLPath \cite{SSLPath}} & \multicolumn{2}{c}{PathDino} \\
    \cline{4-17}
        &&& Yottixel & FPS & Yottixel & FPS & Yottixel & FPS & Yottixel & FPS & Yottixel & FPS & Yottixel & FPS & Yottixel & FPS \\
    \hline\hline
    \multirow{10}{*}{\rotatebox[origin=c]{90}{\textbf{Internal Data}}} & \multirow{1}{*}{Private-Breast} & Top 1 & 24 & \textbf{46}& 33  &\textbf{39} &  39  & \textbf{39}& 45  &\textbf{58} & \textbf{56} &47 &  \textbf{55}  & 51 &60 & \textbf{66} \\\cline{2-17}
    
    & \multirow{3}{*}{Private-Liver} & Top 1 &\textbf{61}  & 49&  \textbf{52}  &46 & \textbf{59}  &50 &  58  &\textbf{60} & 62 &\textbf{66} &   \textbf{65}  & 54&\textbf{76} & 74\\
    & & MV@3 &\textbf{59}  & 42&  \textbf{50}  &47 &    \textbf{68}  & 53& \textbf{68}  &63 & \textbf{67} &62 &  \textbf{69}  &66 &\textbf{83} & 74\\
    & & MV@5 &\textbf{54}  &53 & \textbf{56}  &52 &  \textbf{64}  &56 & 59  & \textbf{67}& \textbf{65} &61 &   \textbf{74}  &62 & \textbf{81}& 74\\\cline{2-17}
    
    & \multirow{3}{*}{Private-Skin} & Top 1&54  &\textbf{65} &  62  &\textbf{66} & 61  & \textbf{63}&  63  &\textbf{65} &  \textbf{70} & 65&  \textbf{71}  & 67 &\textbf{68} &67\\
    & & MV@3& 58 &\textbf{66} &  65  & \textbf{66}& 62  &\textbf{65} &  65  &\textbf{66} &  70  &\textbf{71} &  \textbf{68} & 67&\textbf{69} & 67\\
    & & MV@5 &58  &\textbf{66} & 67  &\textbf{67} &   \textbf{65}  &62 &   67  &\textbf{69} &\textbf{70}  &69 &  66  &\textbf{67}  & 68& \textbf{69}\\\cline{2-17}
    
    & \multirow{3}{*}{Private-CRC} & Top 1 & 51 &\textbf{56} &   54  & \textbf{56}& 54  &\textbf{57} &  \textbf{59}  & 65& 60  & \textbf{64}& 60  & \textbf{64} &58 & \textbf{64}\\
    & & MV@3 &49  & \textbf{59}&   55  &\textbf{58} &  58  & \textbf{62}&  60  &\textbf{68} & 61 &\textbf{66} &  61 & \textbf{65} & 60& \textbf{66}\\
    & & MV@5 & 51 &\textbf{59} &   50  &\textbf{60} & 57  &\textbf{65} & 61  &\textbf{70} & 60 &\textbf{63} &  63  &\textbf{64}  &60 & \textbf{66}\\\hline\hline
    
    \multirow{9}{*}{\rotatebox[origin=c]{90}{\textbf{Public Data}}} & \multirow{3}{*}{PANDA \cite{DataPANDA}}  & Top 1 & \textbf{31} & 28& 32  & \textbf{34}&31   &\textbf{31} & 53  & \textbf{56}& \textbf{59}  & 58&\textbf{47}   & 46 & 59& \textbf{59}\\
    & & MV@3 & \textbf{30} &29 &  33  &\textbf{35} &32   & \textbf{33}& 52  &\textbf{54} & \textbf{58} & 56& \textbf{48}  &45 &58 & \textbf{58}\\
    & & MV@5 &\textbf{31}  & 30& 35   &\textbf{35} & 33  &\textbf{34} & 51  &\textbf{52} & \textbf{55}  &54 & \textbf{48}  &45  & \textbf{57}& 56\\\cline{2-17}

    & \multirow{3}{*}{CAMELYON16 \cite{DataCAMELYON16}} & Top 1 &56  &\textbf{57} & \textbf{65}   &61 & \textbf{59}  &57 & 68  &\textbf{69} &72   &\textbf{74} & 57 & \textbf{70} & \textbf{73}& 69\\
    & & MV@3 & \textbf{56} &50 & \textbf{64}   & 58& 55  & \textbf{61}&68   &\textbf{72} &68   &\textbf{74} &\textbf{69}  & 59 &\textbf{74} & 73\\
    & & MV@5 & 54 & \textbf{54}& \textbf{62}   & 61& \textbf{61}  &59 & 65  &\textbf{70} & 75  &\textbf{77} &\textbf{70}  &61  & \textbf{73}&  72\\\cline{2-17}
    
    & \multirow{3}{*}{BRACS \cite{DataBRACS}} & Top 1 &\textbf{48}  &45 &  47  &\textbf{52} &48   & \textbf{48}& \textbf{56}  & 53& \textbf{59}  &57 & 54  & \textbf{57} & \textbf{59}& 57\\
    & & MV@3 &\textbf{48}  &46 & 51   &\textbf{54}& 49  &\textbf{55} & \textbf{59}  &57 & \textbf{59}  &57 &53  & \textbf{57} & 58& \textbf{58}\\
    & & MV@5&\textbf{50}  &49 & 53   &\textbf{55} & 49  & \textbf{54}& \textbf{60}  &57 &58   & \textbf{59}& 53 &\textbf{57}  &57 & \textbf{59}\\
    \hline\hline
    \end{tabular}
    \end{adjustbox}
    \label{tab:wsiLevelFPSvsYottixelF1}
\end{table*}

\begin{table*}[htbp]
    \centering
\caption{Quantitative Assessment via 5-Fold Cross-Validation: Accuracy in Patch-Level Classification in Histopathological Image Analysis.}
    \begin{adjustbox}{width=1.0\textwidth}
    \small
    \begin{tabular}{ccl|cccc|cccc|cccc}
    \hline\hline
    &&& \multicolumn{4}{c|}{Internal Datasets} & \multicolumn{8}{c}{Public Datasets}     \\\cline{4-15} 
    &&& Private-Breast & Private-Liver &  Private-Skin & Private-CRC & PANDA \cite{DataPANDA} & CAMELYON16 \cite{DataCAMELYON16}& BRACS \cite{DataBRACS}& DigestPath \cite{DataDigestPath}& Kather \cite{DatasetKather}& PanNuke \cite{DataPanNuke}& WSSS4LUAD \cite{DataWSSS4LUAD}\\
    \hline\hline
    \multirow{11}{*}{\rotatebox[origin=c]{90}{Pret. on Natural Data}} & \multirow{6}{*}{\rotatebox[origin=c]{90}{CNN-based}}
    &ResNet50 \cite{Resnet}&62.86$\pm$2.68&	71.77$\pm$3.06&	74.88$\pm$1.23&	53.84$\pm$2.63&	37.22$\pm$0.76&	66.30$\pm$1.88	&54.97$\pm$0.21 &92.18$\pm$2.79	&98.82$\pm$0.36	&77.75$\pm$1.80	&81.79$\pm$0.65& \\
    &&DenseNet121 \cite{DenseNet}&57.75$\pm$1.44&	72.24$\pm$2.06	&72.71$\pm$5.13	&51.22$\pm$2.38	&34.66$\pm$1.61&	65.79$\pm$4.86	&47.92$\pm$4.48&91.85$\pm$3.12&	98.06$\pm$0.24	&73.27$\pm$1.92&	82.35$\pm$0.92 \\
    &&EfficientNet-b3-288 \cite{Efficientnet} &57.99$\pm$2.12&	74.00$\pm$1.13&	74.76$\pm$0.80&	57.52$\pm$1.34&	37.19$\pm$0.59&	63.88$\pm$2.11&	56.51$\pm$0.63&91.25$\pm$2.58&	98.04$\pm$0.33&	72.49$\pm$2.60	&83.78$\pm$0.45 \\
    &&EfficientNet-b5 \cite{Efficientnet}&75.31$\pm$0.74	&78.94$\pm$4.34	&78.67$\pm$2.78	&63.73$\pm$3.06	&40.00$\pm$0.81	&70.16$\pm$5.25	&61.48$\pm$0.76&92.79$\pm$2.68&	99.03$\pm$0.20	&80.29$\pm$1.51&	84.44$\pm$3.05 \\
    &&ConvNext-b-224 \cite{Convnext} &73.24$\pm$1.54&	78.02$\pm$1.20	&77.09$\pm$0.91	&57.33$\pm$2.93	&38.69$\pm$0.97&	69.46$\pm$2.11&	57.87$\pm$1.59&93.88$\pm$1.98&	99.35$\pm$0.11&	82.68$\pm$1.56&	85.18$\pm$0.58 \\
    &&ConvNext-xlarge \cite{Convnext}&76.88$\pm$1.23	&79.22$\pm$0.99&	79.96$\pm$0.86	&60.53$\pm$1.98	&38.91$\pm$1.95	&70.13$\pm$3.64&	57.85$\pm$1.10&94.84$\pm$2.16&	99.72$\pm$0.12&	87.53$\pm$1.76&	87.54$\pm$0.93 \\\cline{2-14}
    & \multirow{5}{*}{\rotatebox[origin=c]{90}{Transformer}}
    &ViT-b16-224 \cite{ViT}&53.27$\pm$3.03	&76.17$\pm$1.82&	75.79$\pm$1.15&	51.58$\pm$4.03&	35.08$\pm$1.06	&64.99$\pm$3.41	&52.03$\pm$1.13&90.76$\pm$2.56&	99.04$\pm$0.36&	79.09$\pm$1.82&	78.26$\pm$8.70 \\
    &&DinoV1-ViT-s16 \cite{DinoV1}&66.85$\pm$2.62	&76.67$\pm$2.60&	70.49$\pm$6.51&	54.00$\pm$2.24&	33.24$\pm$1.40&	69.66$\pm$3.69	&52.91$\pm$2.00&95.47$\pm$1.59	&99.35$\pm$0.22	&82.08$\pm$2.19&	83.87$\pm$2.88 \\
    &&DinoV1-ViT-b16 \cite{DinoV1}&67.54$\pm$1.67&	79.86$\pm$1.83&	73.11$\pm$3.10	&57.93$\pm$2.38&	31.65$\pm$3.64&	69.17$\pm$4.00&	54.70$\pm$4.22&94.75$\pm$1.99&	99.35$\pm$0.18&	85.25$\pm$2.45	&86.63$\pm$1.48 \\
    &&DinoV2-ViT-b14 \cite{DinoV2}&59.91$\pm$2.90	&74.97$\pm$5.76&	71.59$\pm$5.64&	55.18$\pm$2.15	&34.51$\pm$2.55	&64.60$\pm$6.80	&50.75$\pm$4.31&89.88$\pm$6.30&	98.22$\pm$0.72&	70.74$\pm$3.08	&73.19$\pm$8.17 \\
    &&CLIP - ViT-B/16 \cite{CLIP}&55.04$\pm$2.89	&80.55$\pm$1.29&	78.65$\pm$1.87&	59.43$\pm$5.62&	40.44$\pm$0.77&	71.16$\pm$3.47&	59.46$\pm$2.21&88.75$\pm$3.24&	98.65$\pm$0.29&	72.03$\pm$2.77&	82.95$\pm$1.30 \\\hline\hline

    \multirow{12}{*}{\rotatebox[origin=c]{90}{Pret. on Histopathology Data}} 
& \multirow{5}{*}{\rotatebox[origin=c]{90}{CNN-based}}
 & Barlow-Twins-ResNet50  \cite{SSLPath}    &76.15$\pm$3.96&	85.92$\pm$6.54&	82.80$\pm$1.46	&\textbf{71.54$\pm$5.05}	&\textbf{45.50$\pm$1.87}&	76.77$\pm$5.48&\textbf{	66.58$\pm$2.36 }&89.11$\pm$6.83&	\textcolor{gray}{99.53$\pm$0.25}	&\textcolor{gray}{78.14$\pm$3.66}	&\textcolor{gray}{89.16$\pm$0.96} \\
    && MoCoV2-ResNet50 \cite{SSLPath} & 79.59$\pm$1.81&	84.37$\pm$0.62&	79.77$\pm$0.49&	69.40$\pm$0.54&	44.10$\pm$0.35	&74.94$\pm$3.01	&63.83$\pm$0.57&95.88$\pm$1.63&	\textcolor{gray}{96.60$\pm$0.75	}&\textcolor{gray}{61.68$\pm$2.81}	&\textcolor{gray}{83.07$\pm$1.08} \\
    && MuDiPath-ResNet50 \cite{MuDiPath} & 53.08$\pm$6.68&	77.32$\pm$1.80&	74.38$\pm$2.08&	55.29$\pm$2.59&	37.40$\pm$1.00&	65.32$\pm$6.19	&50.87$\pm$3.07&90.49$\pm$1.25	&\textcolor{gray}{98.97$\pm$0.21}&	\textcolor{gray}{79.24$\pm$1.32}&\textcolor{gray}{	83.47$\pm$1.01} \\
    && MuDiPath-DenseNet-101 \cite{MuDiPath} &54.06$\pm$6.78	&74.58$\pm$1.43	&73.97$\pm$2.35&	54.07$\pm$3.09&	34.32$\pm$2.26&	64.65$\pm$3.33	&51.13$\pm$5.84 &91.86$\pm$1.75&\textcolor{gray}{	98.80$\pm$0.26	}&\textcolor{gray}{75.87$\pm$1.28}&\textcolor{gray}{	81.39$\pm$2.15} \\
    && KimiaNet \cite{kimianet}  & 77.03$\pm$4.13&	\textbf{89.25$\pm$1.46}&	\textbf{83.45$\pm$2.00}&	69.58$\pm$3.74&	38.74$\pm$5.74&	79.48$\pm$3.34&	64.75$\pm$3.70&94.58$\pm$2.40&	\textcolor{gray}{99.26$\pm$0.21}&	\textcolor{gray}{85.42$\pm$1.68	}&\textcolor{gray}{82.48$\pm$2.41} \\ \cline{2-14}
& \multirow{6}{*}{\rotatebox[origin=c]{90}{Transformer}}
    & BiomedCLIP \cite{zhang2023large} & 50.76$\pm$3.93&	73.50$\pm$1.34	&75.62$\pm$0.59	&56.13$\pm$2.23&	37.41$\pm$0.31	&67.44$\pm$1.62	&56.53$\pm$0.83&93.79$\pm$1.48&	\textcolor{gray}{96.09$\pm$0.67}&\textcolor{gray}{	64.14$\pm$2.01	}&\textcolor{gray}{84.39$\pm$1.11 }\\
    && HIPT-ViT-s16 \cite{HIPT} &53.07$\pm$6.57	&74.93$\pm$3.75&	74.36$\pm$2.37&	50.53$\pm$2.93	&33.70$\pm$3.27&	65.63$\pm$5.92	&52.05$\pm$5.54 &88.27$\pm$4.73	&\textcolor{gray}{97.58$\pm$0.50}&\textcolor{gray}{	65.52$\pm$3.86}&	\textcolor{gray}{83.49$\pm$1.10} \\
    && TransPath \cite{TransPath} &20.12$\pm$10.81&	61.89$\pm$4.59&	65.22$\pm$1.48&	42.68$\pm$2.57	&29.95$\pm$1.46&	56.80$\pm$10.84	&52.51$\pm$4.60 &71.78$\pm$10.28	&\textcolor{gray}{78.82$\pm$4.92}&\textcolor{gray}{	30.49$\pm$11.68	}&\textcolor{gray}{73.65$\pm$1.93} \\
    && PLIP \cite{PLIP} &56.66$\pm$2.72&	77.03$\pm$1.45	&78.81$\pm$0.56&	64.14$\pm$2.18&	42.16$\pm$0.26&	74.16$\pm$1.04&	61.18$\pm$0.42 &93.63$\pm$2.33&\textcolor{gray}{	92.52$\pm$1.29}&\textcolor{gray}{	60.86$\pm$2.26	}&\textcolor{gray}{77.77$\pm$1.25} \\
    && iBOT-Path  \cite{iBOT} &86.92$\pm$1.97&	86.66$\pm$0.60&	81.51$\pm$0.73&	68.33$\pm$0.59	&40.22$\pm$1.03	&74.96$\pm$1.71	&61.66$\pm$2.85 &\textbf{96.73$\pm$1.13}&\textcolor{gray}{	99.83$\pm$0.12	}&\textcolor{gray}{98.13$\pm$0.57}&	\textcolor{gray}{90.04$\pm$0.97} \\
    && DinoSSLPathology-8 \cite{SSLPath} &81.06$\pm$1.31&	80.45$\pm$4.27&	78.56$\pm$0.85&	59.28$\pm$3.96	&36.82$\pm$0.53	&74.21$\pm$2.95	&58.48$\pm$2.02 &95.53$\pm$1.27	&\textcolor{gray}{99.74$\pm$0.06}&\textcolor{gray}{	92.50$\pm$1.35}	&\textcolor{gray}{89.31$\pm$0.68} \\
\rowcolor{Gray} && PathDino-224 (ours)  & 79.24$\pm$3.84	&81.85$\pm$0.37	&77.32$\pm$1.36	&60.83$\pm$1.98	&33.30$\pm$1.38&	70.03$\pm$5.43&	53.73$\pm$3.72&95.39$\pm$2.25&\textcolor{gray}{	99.72$\pm$0.14}&\textcolor{gray}{	89.63$\pm$1.30	}&\textcolor{gray}{88.34$\pm$1.61 }\\
\rowcolor{Gray} && PathDino-512 (ours)  &\textbf{89.92$\pm$2.65}&	89.16$\pm$2.29&	82.01$\pm$0.90&	70.68$\pm$2.36	&38.47$\pm$2.66&	\textbf{81.42$\pm$1.04}&	63.05$\pm$4.44 &96.52$\pm$1.82&\textcolor{gray}{	99.67$\pm$0.08}&	\textcolor{gray}{91.76$\pm$2.05}	&\textcolor{gray}{88.83$\pm$0.67 }\\ \hline
\hline
    \end{tabular}
    \end{adjustbox}
    \label{tab:crossValidAcc}
\end{table*}

\begin{table*}[htbp]
    \centering
\caption{Quantitative Assessment via 5-Fold Cross-Validation: Macro-Averaged $F$-$1$ Score for Patch-Level Classification in Histopathological Image Analysis.}
    \begin{adjustbox}{width=1.0\textwidth}
    \small
    \begin{tabular}{ccl|cccc|cccc|cccc}
    \hline\hline
    &&& \multicolumn{4}{c|}{Internal Datasets} & \multicolumn{8}{c}{Public Datasets}     \\\cline{4-15} 
    &&& Private-Breast & Private-Liver &  Private-Skin & Private-CRC & PANDA \cite{DataPANDA} & CAMELYON16 \cite{DataCAMELYON16}& BRACS \cite{DataBRACS}& DigestPath \cite{DataDigestPath}& Kather \cite{DatasetKather}& PanNuke \cite{DataPanNuke}& WSSS4LUAD \cite{DataWSSS4LUAD}\\
    \hline\hline
    \multirow{11}{*}{\rotatebox[origin=c]{90}{Pret. on Natural Data}} & \multirow{6}{*}{\rotatebox[origin=c]{90}{CNN-based}}
                      &ResNet50 \cite{Resnet}& 55.80 $\pm$2.19	&54.77 $\pm$4.96	&61.31 $\pm$0.76	&53.69 $\pm$2.98&	27.90 $\pm$1.17&	64.10 $\pm$1.81&	45.96 $\pm$1.80	& 90.13$\pm$2.86&	98.54 $\pm$0.49	&61.37$\pm$2.16	&65.30 $\pm$6.95&	 \\ 
                &&DenseNet121 \cite{DenseNet}& 51.84 $\pm$3.32	&54.63 $\pm$5.78	&58.42 $\pm$1.49&	50.54 $\pm$2.78&	26.14 $\pm$1.47&	62.19 $\pm$3.57&	40.48 $\pm$1.90 &90.31$\pm$4.10	&97.43 $\pm$0.34	&54.56$\pm$1.68	&69.54 $\pm$8.54\\ 
    &&EfficientNet-b3-288 \cite{Efficientnet} &51.17 $\pm$3.01	&61.43 $\pm$2.78	&63.73 $\pm$0.68&	57.58 $\pm$1.35&	27.60 $\pm$0.81&	61.04 $\pm$1.50&	47.85 $\pm$0.88 &89.35$\pm$3.74&	97.63 $\pm$0.24&	53.60$\pm$2.94&	70.04 $\pm$8.42\\ 
        &&EfficientNet-b5 \cite{Efficientnet}&70.41 $\pm$2.86	&72.91 $\pm$4.75	&68.18 $\pm$2.76&	63.12 $\pm$3.57&	30.56 $\pm$2.27&	66.52 $\pm$3.83&	53.19 $\pm$0.86 &91.20$\pm$3.88	&98.86 $\pm$0.30&	65.45$\pm$2.07	&71.69 $\pm$9.74\\ 
             &&ConvNext-b-224 \cite{Convnext} &65.02 $\pm$4.39	&67.42 $\pm$1.35	&67.65 $\pm$0.88	&56.41 $\pm$3.70&	27.65 $\pm$0.76&	65.96 $\pm$1.76&	48.52 $\pm$1.80 &92.69$\pm$2.50	&99.24 $\pm$0.09&	69.52$\pm$2.37	&68.24 $\pm$6.55\\ 
            &&ConvNext-xlarge \cite{Convnext}& 70.72 $\pm$1.75	&70.88 $\pm$5.03	&69.23 $\pm$2.60&	60.48 $\pm$1.81&	30.31 $\pm$1.12&	68.01 $\pm$2.61&	49.44 $\pm$3.14 &93.72$\pm$2.82	&99.69 $\pm$0.15&	76.40$\pm$1.80	&74.33 $\pm$8.44\\\cline{2-14}
    & \multirow{5}{*}{\rotatebox[origin=c]{90}{Transformer}}
    &ViT-b16-224 \cite{ViT}& 43.44 $\pm$2.92	&66.79 $\pm$2.21	&62.84 $\pm$1.52	&50.37 $\pm$5.81	&25.88 $\pm$2.93&	63.11 $\pm$2.14&	43.64 $\pm$2.95 &82.81$\pm$7.25&	98.79 $\pm$0.46	&64.73$\pm$2.94&	58.76 $\pm$19.83\\ 
    &&DinoV1-ViT-s16 \cite{DinoV1}& 60.57 $\pm$2.90&	65.24 $\pm$7.94	&58.52 $\pm$2.51	&51.82 $\pm$4.24	&23.62 $\pm$2.60	&66.32 $\pm$3.93	&39.70 $\pm$4.32 &94.06$\pm$2.12&	99.16 $\pm$0.23	&68.40$\pm$2.28&	67.65 $\pm$7.25\\
    &&DinoV1-ViT-b16 \cite{DinoV1}& 62.80 $\pm$3.65	&71.58 $\pm$2.77	&62.23 $\pm$2.33	&57.44 $\pm$2.85	&24.35 $\pm$2.21	&65.89 $\pm$4.27	&47.74 $\pm$2.11 &93.18$\pm$2.88&	99.18 $\pm$0.20	&72.70$\pm$2.52	&73.69 $\pm$8.64\\
    &&DinoV2-ViT-b14 \cite{DinoV2}&51.16 $\pm$4.28&	63.08 $\pm$6.31	&60.34 $\pm$3.60	&53.99 $\pm$3.26	&25.20 $\pm$3.77	&60.70 $\pm$6.66	&41.46 $\pm$4.62 &86.34$\pm$9.14	&97.61 $\pm$1.17&	50.50$\pm$3.99	&69.20 $\pm$8.31\\
    &&CLIP - ViT-B/16 \cite{CLIP}& 48.56 $\pm$1.72&	61.76 $\pm$8.24&	66.91 $\pm$2.59&	57.13 $\pm$8.27	&40.44 $\pm$0.77&	63.52 $\pm$7.99	&51.87 $\pm$3.84 &86.58 $\pm$3.12&	98.36 $\pm$0.32&	50.72 $\pm$3.62&	68.04 $\pm$8.32\\\hline\hline

    \multirow{14}{*}{\rotatebox[origin=c]{90}{Pret. on Histopathology Data}} & \multirow{6}{*}{\rotatebox[origin=c]{90}{CNN-based}}
    & Barlow-Twins-ResNet50  \cite{SSLPath}    & 70.94 $\pm$6.64&	77.81 $\pm$4.48&	71.87 $\pm$4.22&	70.55 $\pm$5.79&\textbf{37.77  $\pm$3.03}	&	70.64 $\pm$9.57&	59.54 $\pm$2.34& 86.53$\pm$8.46&	\textcolor{gray}{99.47 $\pm$0.25}	&\textcolor{gray}{63.77$\pm$4.32}&	\textcolor{gray}{72.56 $\pm$8.62}\\
    && SwAV-ResNet50  \cite{SSLPath}  & 78.23 $\pm$5.79&	84.78 $\pm$5.13&	78.91 $\pm$1.06&	\textbf{74.65 $\pm$2.62}&40.39 $\pm$1.85	&	78.35 $\pm$5.18&	58.17 $\pm$4.98& 95.87$\pm$1.34&\textcolor{gray}{	98.51 $\pm$0.42}&	\textcolor{gray}{62.85$\pm$3.56}&	\textcolor{gray}{71.43 $\pm$7.35}\\ 
   & & MoCoV2-ResNet50 \cite{SSLPath} & 73.11 $\pm$2.02&	66.87 $\pm$6.62&	65.51 $\pm$0.93&	69.76 $\pm$0.56&33.52 $\pm$0.41	&69.47 $\pm$6.53&	52.28 $\pm$1.76& 94.81$\pm$2.11&\textcolor{gray}{	95.61 $\pm$0.84	}&\textcolor{gray}{34.22$\pm$2.72}&\textcolor{gray}{	65.55 $\pm$6.66}\\ 
   & & MuDiPath-ResNet50 \cite{MuDiPath} &45.79 $\pm$7.27&	61.64 $\pm$5.17&	64.42 $\pm$0.77&	54.42 $\pm$3.25&28.44  $\pm$ 1.92&	63.42 $\pm$4.56&	46.83 $\pm$4.10& 87.30$\pm$4.20&\textcolor{gray}{	98.77 $\pm$0.28	}&\textcolor{gray}{65.47$\pm$2.10}&\textcolor{gray}{	69.75 $\pm$8.82}\\ 
    && MuDiPath-DenseNet-101 \cite{MuDiPath} & 52.62 $\pm$4.34&	58.31 $\pm$3.31&	59.57 $\pm$3.28&	52.94 $\pm$4.08&26.62 $\pm$2.21&62.14 $\pm$3.29&	42.14 $\pm$3.97& 89.34$\pm$2.90	&\textcolor{gray}{98.47 $\pm$0.33}&\textcolor{gray}{	59.79$\pm$2.45}&\textcolor{gray}{	66.46 $\pm$9.08}\\
    &&KimiaNet \cite{kimianet}  & 76.01 $\pm$1.76&	85.77 $\pm$4.22&	76.78 $\pm$1.95&	69.56 $\pm$4.00&	34.09 $\pm$5.32&	77.90 $\pm$2.77&	\textbf{58.35 $\pm$1.35}&93.22$\pm$3.66&\textcolor{gray}{	99.14 $\pm$0.25	}&\textcolor{gray}{72.57$\pm$3.21	}&\textcolor{gray}{65.93 $\pm$10.00}\\ \cline{2-14}
    
& \multirow{8}{*}{\rotatebox[origin=c]{90}{Transformer}}&BiomedCLIP \cite{zhang2023large}
& 38.82 $\pm$1.64&	48.44 $\pm$1.15&	56.62 $\pm$0.61&	55.89 $\pm$2.57& 25.97  $\pm$0.34 &		58.19 $\pm$4.99&	41.89 $\pm$0.93& 92.07$\pm$2.33	&\textcolor{gray}{94.89 $\pm$0.84}&\textcolor{gray}{	37.81$\pm$2.12}&\textcolor{gray}{	69.98 $\pm$8.12}\\ 
&&HIPT-ViT-s16 \cite{HIPT} &43.08 $\pm$6.27&	59.31 $\pm$5.08&	59.47 $\pm$2.85&	48.69 $\pm$4.62&	25.65 $\pm$1.38&	61.56 $\pm$4.25&	42.02 $\pm$8.82& 84.66$\pm$7.17&\textcolor{gray}{	96.81 $\pm$0.69	}&\textcolor{gray}{42.17$\pm$3.80	}&\textcolor{gray}{64.26 $\pm$7.50}\\ 
&&TransPath \cite{TransPath} & 10.17 $\pm$5.14&	39.78 $\pm$5.56&	43.79 $\pm$2.94&	34.81 $\pm$2.74&14.69  $\pm$2.14	&39.23 $\pm$8.06&	38.49 $\pm$0.96&58.28$\pm$12.80&\textcolor{gray}{	72.46$\pm$4.98}&\textcolor{gray}{	12.16$\pm$2.73}&\textcolor{gray}{	50.16 $\pm$10.75}\\ 
&&PLIP \cite{PLIP} &46.07 $\pm$3.20&	50.78 $\pm$1.48&	62.48 $\pm$1.11&	64.11 $\pm$2.70&31.53 $\pm$ 0.47&69.67 $\pm$1.45&	46.72 $\pm$1.05& 92.07$\pm$2.91&\textcolor{gray}{	90.90 $\pm$1.63}&\textcolor{gray}{	27.77$\pm$2.54	}&\textcolor{gray}{61.51 $\pm$7.19}\\ 
&&iBOT-Path  \cite{iBOT} &85.12 $\pm$1.74&	84.37  $\pm$ 1.31	&\textbf{73.09  $\pm$ 0.39}	&68.38   $\pm$ 0.52	&32.95 $\pm$0.86&	73.76  $\pm$ 1.67&	56.52   $\pm$ 1.96& 95.67$\pm$2.17	&\textcolor{gray}{99.81 $\pm$ 0.17	}&\textcolor{gray}{95.76$\pm$1.78	}&\textcolor{gray}{73.31 $\pm$ 5.91}\\ 
&&DinoSSLPathology-8 \cite{SSLPath} &77.59 $\pm$2.17&	74.25 $\pm$4.56&	66.98 $\pm$1.00&	58.17 $\pm$4.77&28.75  $\pm$2.31&70.61 $\pm$1.81&	46.42 $\pm$5.59& 94.50$\pm$1.91&\textcolor{gray}{	99.68 $\pm$0.11}&\textcolor{gray}{	86.17$\pm$2.61}&\textcolor{gray}{	76.30 $\pm$9.60}\\  
\rowcolor{Gray}&&PathDino-224 (ours)  & 78.06 $\pm$4.03&	74.34 $\pm$4.98&	64.89 $\pm$2.14&	60.65 $\pm$2.23&	27.74 $\pm$2.44&	69.26 $\pm$4.94&	46.58 $\pm$3.78&94.03$\pm$3.06&\textcolor{gray}{	99.66  $\pm$ 0.19}&\textcolor{gray}{	81.03$\pm$2.51	}&\textcolor{gray}{74.47 $\pm$ 9.05}\\ 
\rowcolor{Gray}&&PathDino-512 (ours)  &\textbf{88.57 $\pm$3.08}&	\textbf{86.35 $\pm$5.33}&	71.36 $\pm$1.64&	70.47 $\pm$2.47&	32.08 $\pm$2.57&	\textbf{79.61 $\pm$1.00}&	52.59 $\pm$3.21& \textbf{95.82$\pm$2.26}&\textcolor{gray}{	99.65 $\pm$0.11	}&\textcolor{gray}{84.79$\pm$3.14}	&\textcolor{gray}{72.69 $\pm$7.60}\\ \hline\hline

    \end{tabular}
    \end{adjustbox}
    \label{tab:crossValidMacro}
\end{table*}

\begin{table*}[!h]
    \centering
    \caption{Internal histopathology image datasets. Four different datasets were collected at a hospital for four sites including Liver, Skin, Breast, and colon sites.}
    \begin{adjustbox}{width=1.0\textwidth}
    \small
    \begin{tabular}{c|c|c|c|c}
        \hline\hline
        \textbf{Dataset} & \textbf{\#Class} & \textbf{\#WSI} & \textbf{\#Patches} & \textbf{Diagnosis} \\
        \hline\hline
        \multirow{3}{*}{CRC} & \multirow{3}{*}{3} & \multirow{3}{*}{209} & \multirow{3}{*}{4,619} & Cancer Adjacent polyp \\
       &&& & Non-recurrent polyp \\
        &&& & Recurrent polyp \\
        \hline
        \multirow{3}{*}{Liver} & \multirow{3}{*}{3} & \multirow{3}{*}{324} & \multirow{3}{*}{2,976} & Alcoholic Steatohepatitis \\
        &&& & Non-alcoholic Steatohepatitis \\
        &&& & Normal tissue \\
        \hline
        \multirow{4}{*}{Skin} & \multirow{4}{*}{4} & \multirow{4}{*}{660} & \multirow{4}{*}{8,390} & Well differentiated  \\
        &&& & Moderately differentiated \\
        &&& & Poorly differentiated \\
        &&& & Normal tissue \\
        \hline
        \multirow{16}{*}{Breast} & \multirow{16}{*}{16} & \multirow{16}{*}{73} & \multirow{16}{*}{1,141} & Adenoid Cystic Carcinoma \\
        &&& & Adenomyoepthelioma \\
        &&& & Ductal Carcinoma In Situ \\
        &&& & Ductal Carcinoma In Situ, Columnar Cell Lesions Including Flat Epithelial Atypia, Atypical Ductal Hyperplasia \\
       &&& & Intraductal Papilloma, Columnar Cell Lesions \\
       &&& & Invasive Breast Carcinoma of No Special Type \\
       &&& & Invasive lobular carcinoma \\
        &&& & Lobular Carcinoma In Situ + Atypical Lobular Hyperplasia \\
        &&& & Lobular Carcinoma In Situ, Flat Epithelial Atypia, Atypical Lobular Hyperplasia \\
       &&& & Malignant Adenomyoepithelioma \\
        &&& & Metaplastic Carcinoma \\
        &&& & Microglandular Adenosis \\
       &&& & Microinvasive carcinoma \\
        &&& & Mucinous Cystadenocarcinoma \\
       && & & Radial scar complex sclerosing lesion \\
        &&& & Normal tissue\\
        \hline\hline
    \end{tabular}
    \end{adjustbox}
    \label{tab:privateDataset} 
\end{table*}

\begin{table*}[!h]
    \centering
    \caption{Public histopathology image datasets including PANDA, CAMELYON16, BRACS, DigestPath, Kather, PanNuke, and WSSS4LUAD. Number of images represents the total images (patches) used in the evaluation regardless of their training/testing split, since we used the leave-one-out evaluation method for the search task and k-fold cross-validation for the classification task.}
    \begin{adjustbox}{width=0.8\textwidth}
    \small
    \begin{tabular}{c|c|c|c|c|c}
        \hline
        \hline
        \textbf{Dataset} & \textbf{Analysis Scale} & \textbf{\#Class} & \textbf{\#WSI} & \textbf{\#Image} & \textbf{Diagnosis} \\
        \hline\hline
        \multirow{6}{*}{PANDA} & \multirow{6}{*}{WSI/Patch} & \multirow{6}{*}{6} & \multirow{6}{*}{10,349} & \multirow{6}{*}{87,451} 
        & background (non tissue) or unknown
 \\
        &&&&& benign tissue (stroma and epithelium combined)
 \\
        &&&&& cancerous tissue, not specified
 \\
        &&&&& cancerous epithelium (Gleason 3)
 \\
         &&&&& cancerous epithelium (Gleason 4)
 \\
        &&&&& cancerous epithelium (Gleason 5)
 \\
        \hline
        \multirow{2}{*}{CAMELYON16} & \multirow{2}{*}{WSI/Patch} & \multirow{2}{*}{2} & \multirow{2}{*}{128} & \multirow{2}{*}{2,864} & Tumor \\
        &&&&& Normal \\
        \hline
        \multirow{3}{*}{BRACS} & \multirow{3}{*}{WSI/Patch} & \multirow{3}{*}{3} & \multirow{3}{*}{523} & \multirow{3}{*}{10,984} & Benign \\
        &&&&& Atypical \\
        &&&&& Malignant \\
        \hline
        \multirow{2}{*}{DigestPath} & \multirow{2}{*}{Patch-Level} & \multirow{2}{*}{2} & \multirow{2}{*}{-} & \multirow{2}{*}{1,103} & Benign \\
        &&&&& Malignant \\
        \hline
        \multirow{9}{*}{Kather} & \multirow{9}{*}{Patch-Level} & \multirow{9}{*}{9} & \multirow{9}{*}{-} & \multirow{9}{*}{7,180} & ADIPOSE \\
        &&&&& BACKGROUND \\
        &&&&& DEBRIS \\
        &&&&& LYMPHO \\
        &&&&& MUCUS \\
        &&&&& Smooth Muscle \\
        &&&&& Normal Colon Mucosa \\
        &&&&& Cancer-Associated Stroma \\
        &&&&& Colorectal Adenocarcinoma Epithelium \\
        \hline
        \multirow{19}{*}{PanNuke} & \multirow{19}{*}{Patch-Level} & \multirow{19}{*}{19} & \multirow{19}{*}{-} & \multirow{19}{*}{\begin{tabular}[c]{@{}l@{}}2,656\\ 2,523\\ 2,722\end{tabular}} & Breast \\
        &&&&& Colon \\ 
        &&&&& Bile-duct \\
        &&&&& Esophagus \\
        &&&&& Uterus \\
        &&&&& Lung \\
        &&&&& Cervix \\
        &&&&& Head Neck  \\
        &&&&& Skin \\
        &&&&& Adrenal\_gland \\
        &&&&& kidney \\
        &&&&& Stomach \\
        &&&&& Prostate \\
        &&&&& testis \\
        &&&&& Liver \\
        &&&&& Thyroid \\
        &&&&& Pancreatic \\
        &&&&& Overin \\
        &&&&& Bladder \\
        \hline
        \multirow{7}{*}{WSSS4LUAD} & \multirow{7}{*}{Patch-Level} & \multirow{7}{*}{7} & \multirow{7}{*}{-} & \multirow{7}{*}{10,091} & Normal\\
        &&&&& Stroma \\
        &&&&& Stroma-Normal \\
        &&&&& Tumor \\
        &&&&& Tumor-Normal \\
        &&&&& Tumor-Stroma \\
        &&&&& Tumor-Stroma-Normal \\
        \hline\hline
    \end{tabular}
    \end{adjustbox}
     \label{tab:publicDataset}
\end{table*}


\begin{table*}[htbp]
    \centering
\caption{WSI-level  Top-1 Accuracy using the proposed FPS patching method and minimum of medium proposed in Yottixel \cite{yottixel}}
    \begin{adjustbox}{width=1.0\textwidth}
    \small
    \begin{tabular}{ccl|cccc|cccc}
    \hline\hline
    &&& \multicolumn{4}{c|}{Internal Datasets} & \multicolumn{3}{c}{Public Datasets} \\\cline{4-11}
    &&& Private-Breast & Private-Liver &  Private-Skin & Private-CRC & PANDA \cite{DataPANDA}& CAMELYON16 \cite{DataCAMELYON16}& BRACS  \cite{DataBRACS}\\
    \hline\hline
    \multirow{11}{*}{\rotatebox[origin=c]{90}{Pret. on Natural}} & \multirow{6}{*}{\rotatebox[origin=c]{90}{CNN-based}}
    &ResNet50 \cite{Resnet}&0.48&0.67&0.73&0.58&0.32&0.54&0.53&   \\ 
    &&DenseNet121 \cite{DenseNet}&  0.48&0.64&0.69&0.49&0.3&0.67&0.52   \\ 
    &&EfficientNet-b3-288 \cite{Efficientnet}&  0.41&0.66&0.73&0.6&0.32&0.59&0.55  \\ 
    &&EfficientNet-b5 \cite{Efficientnet} &  0.51&0.71&0.71&0.63&0.37&0.57&0.54   \\ 
    &&ConvNext-b-224 \cite{Convnext}& 0.56&0.75&0.74&0.6&0.34&0.62&0.58     \\ 
    &&ConvNext-xlarge  \cite{Convnext}&  0.56&0.76&0.74&0.58&0.35&0.61&0.58     \\\cline{2-11}
    & \multirow{5}{*}{\rotatebox[origin=c]{90}{Transformer}}
    &ViT-b16-224 \cite{ViT}& 0.41&0.7&0.72&0.5&0.31&0.6&0.54   \\ 
    &&DinoV1-ViT-s16 \cite{DinoV1}&  0.48&0.71&0.74&0.55&0.36&0.67&0.6     \\
    &&DinoV1-ViT-b16  \cite{DinoV1}&  0.55&0.72&0.73&0.61&0.37&0.63&0.59    \\
    &&DinoV2-ViT-b14  \cite{DinoV2}&  0.53&0.71&0.72&0.56&0.31&0.61&0.51     \\
    &&CLIP - ViT-B/16 \cite{CLIP}& 0.49&0.67&0.75&0.56&0.36&0.67&0.58     \\\hline\hline
    \multirow{14}{*}{\rotatebox[origin=c]{90}{Pret. on Histopathology}} & \multirow{5}{*}{\rotatebox[origin=c]{90}{CNN-based}}
    & Barlow-Twins-ResNet50 \cite{SSLPath}& 0.58&0.77&0.77&0.64&0.61&0.67&0.61    \\ 
    && MoCoV2-ResNet50 \cite{SSLPath}&  0.62&0.79&0.74&0.64&0.62&0.67&0.6   \\ 
    && MuDiPath-ResNet50 \cite{MuDiPath}& 0.44&0.7&0.72&0.58&0.35&0.63&0.51    \\ 
    && MuDiPath-DenseNet-101  \cite{MuDiPath}&  0.51&0.68&0.74&0.66&0.36&0.65&0.56  \\
    && KimiaNet \cite{kimianet} &  0.51&0.78&0.75&0.64&0.57&0.76&0.62    \\\cline{2-11}
    & \multirow{6}{*}{\rotatebox[origin=c]{90}{Transformer}}
    & BiomedCLIP - \cite{zhang2023large} & 0.47&0.74&0.75&0.58&0.34&0.61&0.55     \\
    && HIPT-ViT-s16 \cite{HIPT} &  0.44&0.68&0.73&0.57&0.32&0.62&0.52    \\ 
    && PLIP \cite{PLIP} &  0.58&0.73&0.75&0.64&0.56&0.73&0.6    \\ 
    && iBOT-Path  \cite{iBOT} & 0.64&0.79&0.76&0.65&0.53&0.67&0.64     \\ 
    && DinoSSLPathology-8 \cite{SSLPath} &  0.58&0.74&0.78&0.63&0.47&0.74&0.61  \\
    && PathDino-224 (ours)&  0.53&0.75&0.74&0.61&0.46&0.72&0.61   \\
    && PathDino-512 (ours)& 0.68&0.83&0.78&0.63&0.58&0.73&0.64   \\\hline\hline
    \end{tabular}
    \end{adjustbox}
    \label{tab:WSIAccTop1}
\end{table*}

\begin{table*}[htbp]
    \centering
\caption{WSI-level  Top-1 Macro avg $F$-$1$ score}
    \begin{adjustbox}{width=1.0\textwidth}
    \small
    \begin{tabular}{ccl|cccc|cccc}
    \hline\hline
    &&& \multicolumn{4}{c|}{Internal Datasets} & \multicolumn{3}{c}{Public Datasets} \\\cline{4-11}
    &&& Private-Breast & Private-Liver &  Private-Skin & Private-CRC & PANDA \cite{DataPANDA}& CAMELYON16 \cite{DataCAMELYON16}& BRACS  \cite{DataBRACS}\\
    \hline\hline
    \multirow{11}{*}{\rotatebox[origin=c]{90}{Pret. on Natural}} & \multirow{6}{*}{\rotatebox[origin=c]{90}{CNN-based}}
    &ResNet50 \cite{Resnet}&  0.43&0.49&0.63&0.58&0.29&0.48&0.48  & \\ 
    &&DenseNet121 \cite{DenseNet}& 0.37&0.44&0.61&0.48&0.27&0.65&0.47  \\ 
    &&EfficientNet-b3-288 \cite{Efficientnet}& 0.35&0.49&0.64&0.6&0.29&0.55&0.5  \\ 
    &&EfficientNet-b5 \cite{Efficientnet} &  0.41&0.52&0.6&0.63&0.35&0.54&0.49   \\ 
    &&ConvNext-b-224 \cite{Convnext}& 0.47&0.61&0.64&0.6&0.31&0.6&0.53      \\ 
    &&ConvNext-xlarge  \cite{Convnext}&  0.51&0.52&0.64&0.58&0.33&0.57&0.52      \\\cline{2-11}
    & \multirow{5}{*}{\rotatebox[origin=c]{90}{Transformer}}
    &ViT-b16-224 \cite{ViT}&  0.3&0.51&0.62&0.5&0.28&0.54&0.49    \\ 
    &&DinoV1-ViT-s16 \cite{DinoV1}&  0.41&0.52&0.64&0.55&0.34&0.63&0.55     \\
    &&DinoV1-ViT-b16  \cite{DinoV1}&  0.49&0.59&0.62&0.61&0.35&0.6&0.54    \\
    &&DinoV2-ViT-b14  \cite{DinoV2}&  0.46&0.49&0.65&0.56&0.28&0.57&0.45    \\
    &&CLIP - ViT-B/16 \cite{CLIP}& 0.39&0.46&0.66&0.56&0.34&0.61&0.52    \\\hline\hline
    \multirow{14}{*}{\rotatebox[origin=c]{90}{Pret. on Histopathology}} & \multirow{5}{*}{\rotatebox[origin=c]{90}{CNN-based}}
    & Barlow-Twins-ResNet50 \cite{SSLPath}& 0.49&0.56&0.65&0.65&0.62&0.63&0.56     \\
    && MoCoV2-ResNet50 \cite{SSLPath}&  0.49&0.63&0.63&0.64&0.64&0.64&0.53   \\ 
    && MuDiPath-ResNet50 \cite{MuDiPath}& 0.41&0.57&0.61&0.58&0.33&0.59&0.45    \\ 
    && MuDiPath-DenseNet-101  \cite{MuDiPath}& 0.47&0.5&0.63&0.66&0.33&0.63&0.5  \\
    && KimiaNet \cite{kimianet} & 0.47&0.66&0.65&0.64&0.58&0.74&0.57   \\\cline{2-11}
    & \multirow{6}{*}{\rotatebox[origin=c]{90}{Transformer}}
    & BiomedCLIP - \cite{zhang2023large} &  0.39&0.5&0.63&0.57&0.31&0.57&0.48     \\
    && HIPT-ViT-s16 \cite{HIPT} &  0.33&0.46&0.62&0.57&0.29&0.58&0.48    \\ 
    && PLIP \cite{PLIP} &  0.58&0.6&0.65&0.65&0.56&0.69&0.53  \\ 
    && iBOT-Path  \cite{iBOT} &  0.61&0.74&0.66&0.66&0.52&0.62&0.58     \\ 
    && DinoSSLPathology-8 \cite{SSLPath} & 0.51&0.54&0.67&0.64&0.46&0.7&0.57  \\
    && PathDino-224 (ours)& 0.56&0.6&0.63&0.61&0.45&0.69&0.56   \\
    && PathDino-512 (ours)& 0.66&0.74&0.67&0.64&0.59&0.69&0.57   \\\hline\hline
    \end{tabular}
    \end{adjustbox}
    \label{tab:WSIMacroTop1}
\end{table*}

\begin{table*}[htbp]
    \centering
\caption{WSI-level  MV@3 Accuracy}
    \begin{adjustbox}{width=1.0\textwidth}
    \small
    \begin{tabular}{ccl|ccc|cccc}
    \hline\hline
    &&& \multicolumn{3}{c|}{Internal Datasets} & \multicolumn{3}{c}{Public Datasets} \\\cline{4-10}
    &&&  Private-Liver &  Private-Skin & Private-CRC & PANDA \cite{DataPANDA}& CAMELYON16 \cite{DataCAMELYON16}& BRACS  \cite{DataBRACS}\\
    \hline\hline
    \multirow{11}{*}{\rotatebox[origin=c]{90}{Pret. on Natural}} & \multirow{6}{*}{\rotatebox[origin=c]{90}{CNN-based}}
    &ResNet50 \cite{Resnet}&0.72&0.76&0.64&0.34&0.6&0.58&    \\ 
    &&DenseNet121 \cite{DenseNet}&0.72&0.74&0.51&0.32&0.67&0.54    \\ 
    &&EfficientNet-b3-288 \cite{Efficientnet}&0.69&0.76&0.61&0.33&0.6&0.58   \\ 
    &&EfficientNet-b5 \cite{Efficientnet} &0.71&0.75&0.64&0.38&0.62&0.57      \\ 
    &&ConvNext-b-224 \cite{Convnext}&0.75&0.78&0.62&0.37&0.68&0.6       \\ 
    &&ConvNext-xlarge  \cite{Convnext}&0.76&0.79&0.61&0.37&0.65&0.62       \\\cline{2-10}
    & \multirow{5}{*}{\rotatebox[origin=c]{90}{Transformer}}
    &ViT-b16-224 \cite{ViT}&0.71&0.76&0.55&0.33&0.67&0.57      \\ 
    &&DinoV1-ViT-s16 \cite{DinoV1}&0.74&0.77&0.61&0.38&0.67&0.59      \\
    &&DinoV1-ViT-b16  \cite{DinoV1}&0.77&0.77&0.62&0.39&0.69&0.63      \\
    &&DinoV2-ViT-b14  \cite{DinoV2}&0.74&0.77&0.59&0.32&0.58&0.53       \\
    &&CLIP - ViT-B/16 \cite{CLIP}& 0.69&0.79&0.58&0.38&0.67&0.6      \\\hline\hline
    \multirow{14}{*}{\rotatebox[origin=c]{90}{Pret. on Histopathology}} & \multirow{5}{*}{\rotatebox[origin=c]{90}{CNN-based}}
    & Barlow-Twins-ResNet50 \cite{SSLPath}& 0.79&0.77&0.66&0.58&0.7&0.64       \\
    && MoCoV2-ResNet50 \cite{SSLPath}&  0.83&0.78&0.65&0.6&0.69&0.61     \\ 
    && MuDiPath-ResNet50 \cite{MuDiPath}& 0.73&0.78&0.58&0.37&0.67&0.56       \\ 
    && MuDiPath-DenseNet-101  \cite{MuDiPath}&0.72&0.77&0.63&0.37&0.68&0.58  \\
    && KimiaNet \cite{kimianet}6&0.8&0.81&0.65&0.56&0.78&0.64     \\\cline{2-10}
    & \multirow{6}{*}{\rotatebox[origin=c]{90}{Transformer}}
    & BiomedCLIP - \cite{zhang2023large} &0.77&0.78&0.63&0.36&0.67&0.62     \\
    && HIPT-ViT-s16 \cite{HIPT} &  0.67&0.76&0.52&0.33&0.64&0.54      \\ 
    && PLIP \cite{PLIP} &  0.74&0.79&0.67&0.55&0.77&0.63     \\ 
    && iBOT-Path  \cite{iBOT} & 0.83&0.8&0.63&0.53&0.71&0.65    \\ 
    && DinoSSLPathology-8 \cite{SSLPath} &0.77&0.8&0.64&0.47&0.68&0.64  \\
    && PathDino-224 (ours)&  0.79&0.8&0.59&0.48&0.74&0.61      \\
    && PathDino-512 (ours)& 0.86&0.8&0.65&0.58&0.78&0.66      \\\hline\hline
    \end{tabular}
    \end{adjustbox}
    \label{tab:WSIAccMV3}
\end{table*}

\begin{table*}[htbp]
    \centering
\caption{WSI-level  MV@3 Macro Avg $F$-$1$ score}
    \begin{adjustbox}{width=1.0\textwidth}
    \small
    \begin{tabular}{ccl|ccc|cccc}
    \hline\hline
    &&& \multicolumn{3}{c|}{Internal Datasets} & \multicolumn{3}{c}{Public Datasets} \\\cline{4-10}
    &&&  Private-Liver &  Private-Skin & Private-CRC & PANDA \cite{DataPANDA}& CAMELYON16 \cite{DataCAMELYON16}& BRACS  \cite{DataBRACS}\\
    \hline\hline
    \multirow{11}{*}{\rotatebox[origin=c]{90}{Pret. on Natural}} & \multirow{6}{*}{\rotatebox[origin=c]{90}{CNN-based}}
    &ResNet50 \cite{Resnet}&0.49&0.65&0.64&0.3&0.52&0.53&    \\ 
    &&DenseNet121 \cite{DenseNet}& 0.49&0.62&0.5&0.28&0.62&0.48     \\ 
    &&EfficientNet-b3-288 \cite{Efficientnet}&   0.5&0.65&0.6&0.3&0.52&0.52   \\ 
    &&EfficientNet-b5 \cite{Efficientnet} &  0.55&0.61&0.64&0.35&0.56&0.53     \\ 
    &&ConvNext-b-224 \cite{Convnext}& 0.6&0.67&0.61&0.33&0.63&0.53      \\ 
    &&ConvNext-xlarge  \cite{Convnext}& 0.56&0.66&0.61&0.34&0.6&0.54       \\\cline{2-10}
    & \multirow{5}{*}{\rotatebox[origin=c]{90}{Transformer}}
    &ViT-b16-224 \cite{ViT} &0.48&0.64&0.54&0.29&0.61&0.51     \\ 
    &&DinoV1-ViT-s16 \cite{DinoV1}&   0.54&0.65&0.61&0.34&0.6&0.53     \\
    &&DinoV1-ViT-b16  \cite{DinoV1}&  0.66&0.65&0.62&0.35&0.64&0.58      \\
    &&DinoV2-ViT-b14  \cite{DinoV2}& 0.54&0.66&0.59&0.29&0.5&0.46      \\
    &&CLIP - ViT-B/16 \cite{CLIP}&0.47&0.66&0.58&0.35&0.58&0.54      \\\hline\hline
    \multirow{14}{*}{\rotatebox[origin=c]{90}{Pret. on Histopathology}} & \multirow{5}{*}{\rotatebox[origin=c]{90}{CNN-based}}
    & Barlow-Twins-ResNet50 \cite{SSLPath}&0.63&0.62&0.67&0.59&0.64&0.57      \\
    && MoCoV2-ResNet50 \cite{SSLPath}& 0.71&0.64&0.65&0.6&0.65&0.54     \\ 
    && MuDiPath-ResNet50 \cite{MuDiPath}&0.5&0.64&0.58&0.33&0.62&0.49      \\ 
    && MuDiPath-DenseNet-101  \cite{MuDiPath}& 0.53&0.63&0.63&0.34&0.64&0.52 \\
    && KimiaNet \cite{kimianet} & 0.62&0.71&0.66&0.56&0.74&0.57     \\\cline{2-10}
    & \multirow{6}{*}{\rotatebox[origin=c]{90}{Transformer}}
    & BiomedCLIP - \cite{zhang2023large} & 0.53&0.65&0.62&0.33&0.61&0.55       \\
    && HIPT-ViT-s16 \cite{HIPT} & 0.45&0.63&0.52&0.3&0.56&0.47     \\ 
    && PLIP \cite{PLIP} & 0.63&0.66&0.68&0.54&0.72&0.57    \\ 
    && iBOT-Path  \cite{iBOT} &0.76&0.69&0.64&0.52&0.64&0.59     \\ 
    && DinoSSLPathology-8 \cite{SSLPath} & 0.66&0.67&0.65&0.45&0.59&0.57 \\
    && PathDino-224 (ours)& 0.6&0.68&0.59&0.46&0.69&0.53      \\
    && PathDino-512 (ours)&0.74&0.67&0.66&0.58&0.73&0.58      \\\hline\hline
    \end{tabular}
    \end{adjustbox}
    \label{tab:WSIMacroMV3}
\end{table*}

\begin{table*}[htbp]
    \centering
\caption{WSI-level  MV@5 Accuracy}
    \begin{adjustbox}{width=1.0\textwidth}
    \small
    \begin{tabular}{ccl|ccc|cccc}
    \hline\hline
    &&& \multicolumn{3}{c|}{Internal Datasets} & \multicolumn{3}{c}{Public Datasets} \\\cline{4-10}
    &&&  Private-Liver &  Private-Skin & Private-CRC & PANDA \cite{DataPANDA}& CAMELYON16 \cite{DataCAMELYON16}& BRACS  \cite{DataBRACS}\\
    \hline\hline
    \multirow{11}{*}{\rotatebox[origin=c]{90}{Pret. on Natural}} & \multirow{6}{*}{\rotatebox[origin=c]{90}{CNN-based}}
    &ResNet50 \cite{Resnet}& 0.71&0.78&0.63&0.36&0.64&0.6   &  \\ 
    &&DenseNet121 \cite{DenseNet}&  0.7&0.78&0.49&0.34&0.7&0.58  \\ 
    &&EfficientNet-b3-288 \cite{Efficientnet}&  0.68&0.78&0.59&0.36&0.61&0.57   \\ 
    &&EfficientNet-b5 \cite{Efficientnet} &0.7&0.77&0.64&0.4&0.66&0.61    \\ 
    &&ConvNext-b-224 \cite{Convnext}&  0.73&0.79&0.58&0.38&0.71&0.62      \\ 
    &&ConvNext-xlarge  \cite{Convnext}&  0.78&0.8&0.63&0.39&0.67&0.64     \\\cline{2-10}
    & \multirow{5}{*}{\rotatebox[origin=c]{90}{Transformer}}
    &ViT-b16-224 \cite{ViT}&  0.72&0.78&0.54&0.35&0.69&0.58     \\ 
    &&DinoV1-ViT-s16 \cite{DinoV1}& 0.76&0.78&0.6&0.39&0.67&0.63     \\
    &&DinoV1-ViT-b16  \cite{DinoV1}&  0.78&0.79&0.62&0.41&0.71&0.64     \\
    &&DinoV2-ViT-b14  \cite{DinoV2}& 0.72&0.78&0.59&0.35&0.63&0.56       \\
    &&CLIP - ViT-B/16 \cite{CLIP}& 0.71&0.8&0.6&0.4&0.69&0.61      \\\hline\hline
    \multirow{14}{*}{\rotatebox[origin=c]{90}{Pret. on Histopathology}} & \multirow{5}{*}{\rotatebox[origin=c]{90}{CNN-based}}
    & Barlow-Twins-ResNet50 \cite{SSLPath}&  0.79&0.79&0.64&0.57&0.74&0.66      \\
    && MoCoV2-ResNet50 \cite{SSLPath}&  0.82&0.78&0.65&0.58&0.71&0.64     \\ 
    && MuDiPath-ResNet50 \cite{MuDiPath}& 0.73&0.79&0.57&0.39&0.7&0.56      \\ 
    && MuDiPath-DenseNet-101  \cite{MuDiPath}& 0.72&0.78&0.66&0.39&0.71&0.6 \\
    && KimiaNet \cite{kimianet} & 0.78&0.82&0.62&0.54&0.81&0.66    \\\cline{2-10}
    & \multirow{6}{*}{\rotatebox[origin=c]{90}{Transformer}}
    & BiomedCLIP - \cite{zhang2023large} & 0.77&0.78&0.65&0.38&0.69&0.61      \\
    && HIPT-ViT-s16 \cite{HIPT} & 0.68&0.77&0.55&0.35&0.67&0.56      \\ 
    && PLIP \cite{PLIP} & 0.76&0.82&0.69&0.54&0.75&0.64      \\ 
    && iBOT-Path  \cite{iBOT} &  0.82&0.82&0.63&0.53&0.74&0.65    \\ 
    && DinoSSLPathology-8 \cite{SSLPath} & 0.77&0.8&0.63&0.48&0.7&0.64 \\
    && PathDino-224 (ours)& 0.76&0.8&0.57&0.48&0.71&0.64     \\
    && PathDino-512 (ours)& 0.85&0.82&0.65&0.56&0.77&0.67      \\\hline\hline
    \end{tabular}
    \end{adjustbox}
    \label{tab:WSIAccMV5}
\end{table*}

\begin{table*}[htbp]
    \centering
\caption{WSI-level  MV@5 Macro Avg $F$-$1$ score}
    \begin{adjustbox}{width=1.0\textwidth}
    \small
    \begin{tabular}{ccl|ccc|cccc}
    \hline\hline
    &&& \multicolumn{3}{c|}{Internal Datasets} & \multicolumn{3}{c}{Public Datasets} \\\cline{4-10}
    &&&  Private-Liver &  Private-Skin & Private-CRC & PANDA \cite{DataPANDA}& CAMELYON16 \cite{DataCAMELYON16}& BRACS  \cite{DataBRACS}\\
    \hline\hline
    \multirow{11}{*}{\rotatebox[origin=c]{90}{Pret. on Natural}} & \multirow{6}{*}{\rotatebox[origin=c]{90}{CNN-based}}
    &ResNet50 \cite{Resnet}&  0.48&0.66&0.63&0.32&0.55&0.54  & \\ 
    &&DenseNet121 \cite{DenseNet}& 0.47&0.65&0.49&0.3&0.64&0.52   \\ 
    &&EfficientNet-b3-288 \cite{Efficientnet}& 0.5&0.65&0.59&0.31&0.49&0.51   \\ 
    &&EfficientNet-b5 \cite{Efficientnet} & 0.47&0.63&0.64&0.36&0.58&0.56      \\ 
    &&ConvNext-b-224 \cite{Convnext}& 0.5&0.65&0.57&0.33&0.65&0.55       \\ 
    &&ConvNext-xlarge  \cite{Convnext}& 0.53&0.66&0.63&0.35&0.57&0.56       \\\cline{2-10}
    & \multirow{5}{*}{\rotatebox[origin=c]{90}{Transformer}}
    &ViT-b16-224 \cite{ViT}& 0.49&0.65&0.53&0.31&0.59&0.5     \\ 
    &&DinoV1-ViT-s16 \cite{DinoV1}& 0.52&0.65&0.61&0.36&0.57&0.56      \\
    &&DinoV1-ViT-b16  \cite{DinoV1}& 0.57&0.66&0.61&0.37&0.65&0.57     \\
    &&DinoV2-ViT-b14  \cite{DinoV2}& 0.53&0.66&0.59&0.3&0.54&0.49       \\
    &&CLIP - ViT-B/16 \cite{CLIP}& 0.52&0.67&0.6&0.35&0.61&0.55      \\\hline\hline
    \multirow{14}{*}{\rotatebox[origin=c]{90}{Pret. on Histopathology}} & \multirow{5}{*}{\rotatebox[origin=c]{90}{CNN-based}}
    & Barlow-Twins-ResNet50 \cite{SSLPath}& 0.67&0.63&0.65&0.56&0.69&0.57       \\
    && MoCoV2-ResNet50 \cite{SSLPath}& 0.66&0.61&0.66&0.57&0.65&0.56     \\ 
    && MuDiPath-ResNet50 \cite{MuDiPath}& 0.49&0.64&0.57&0.35&0.62&0.49      \\ 
    && MuDiPath-DenseNet-101  \cite{MuDiPath}& 0.49&0.63&0.66&0.35&0.66&0.52  \\
    && KimiaNet \cite{kimianet} & 0.61&0.69&0.63&0.54&0.77&0.59      \\\cline{2-10}
    & \multirow{6}{*}{\rotatebox[origin=c]{90}{Transformer}}
    & BiomedCLIP - \cite{zhang2023large} & 0.56&0.62&0.65&0.34&0.59&0.54       \\
    && HIPT-ViT-s16 \cite{HIPT} & 0.46&0.65&0.56&0.3&0.58&0.48       \\ 
    && PLIP \cite{PLIP} & 0.67&0.69&0.7&0.52&0.7&0.57     \\ 
    && iBOT-Path  \cite{iBOT}  &0.72&0.70&0.64&0.51&0.67&0.57     \\ 
    && DinoSSLPathology-8 \cite{SSLPath} & 0.62&0.67&0.64&0.45&0.61&0.57  \\
    && PathDino-224 (ours)&  0.65&0.67&0.58&0.45&0.64&0.56      \\
    && PathDino-512 (ours)& 0.74&0.69&0.66&0.56&0.72&0.59     \\\hline\hline
    \end{tabular}
    \end{adjustbox}
    \label{tab:WSIMacroMV5}
\end{table*}

\begin{algorithm*}
\caption{HistoRotate, Image Augmentation in a Self-Supervised Manner (DINO Framework)}
\label{alg:rotation-agnostic-pipeline}
\begin{algorithmic}[1]
\Require Input image $I$, Global crop scales $[a, b]$, Local crop scales $[c, d]$, Number of local crops $n$, Set of discrete angles $\Theta$

\Function{ExactRotation}{$I, \Theta$}
    \State $\theta \gets \text{random.choice}(\Theta)$
    \State $I' \gets \text{rotate}(I, \theta)$
    \State \Return $I'$
\EndFunction

\Function{HistoRotate}{$I, [a, b], [c, d], n, \Theta$}
    \State Initialize empty list $crops$
    \If{$\text{size}(I)[0] = 1024$}
        \State $crops.\text{append}(\text{global-transfo1-1024}(I))$ \Comment{Include Random $360^\circ$ Rotation}
        \State $crops.\text{append}(\text{global-transfo2-1024}(I))$ \Comment{Include Random $360^\circ$ Rotation}
        \For{$i = 1, n$}
            \State $crops.\text{append}(\text{local-transfo}(I))$ \Comment{Always Include Random $360^\circ$ Rotation}
        \EndFor
    \Else
        \State $crops.\text{append}(\text{global-transfo1\_512}(I))$ \Comment{Include Random Rotation from $\Theta$ = \{90, 180, 270, 360\}}
        \State $crops.\text{append}(\text{global-transfo2\_512}(I))$ \Comment{Include Random Rotation from $\Theta$ = \{90, 180, 270, 360\}}
        \For{$i = 1, n$}
            \State $crops.\text{append}(\text{local-transfo}(I))$ \Comment{Always Include Random $360^\circ$ Rotation}
        \EndFor
    \EndIf
    \State \Return $crops$
\EndFunction
\end{algorithmic}
\end{algorithm*}


\end{document}